\pdfoutput=1

\documentclass[11pt]{article}

\usepackage[final]{acl}

\usepackage{times}
\usepackage{latexsym}
\usepackage{mathtools}

\usepackage[T1]{fontenc}

\usepackage[utf8]{inputenc}

\usepackage{microtype}
\usepackage{subcaption}

\usepackage{inconsolata}
\usepackage{listings}
\usepackage{graphicx}
\usepackage{multirow}
\usepackage{algorithm}
\usepackage{booktabs}
\usepackage{algpseudocode}
\usepackage{diagbox}
\usepackage{tcolorbox}
\newcommand{\xhdr}[1]{{\noindent\bfseries #1}.} 

\title{ConsistencyChecker: Tree-based Evaluation of\\ LLM Generalization Capabilities}

\author{
  Zhaochen Hong\textsuperscript{1*} \ 
  Haofei Yu\textsuperscript{1*} \
  Jiaxuan You\textsuperscript{1} \\
  \textsuperscript{1}University of Illinois Urbana-Champaign \\
  \texttt{\{zhong42,haofeiy2,jiaxuan\}@illinois.edu}
}

\usepackage{amsmath}
\usepackage{amsfonts}

\begin{document}
\maketitle

\begingroup

\renewcommand{\thefootnote}{\fnsymbol{footnote}}
\footnotetext[1]{Two authors contribute equally to this paper.}
\endgroup
\begin{abstract}
Evaluating consistency in Large Language Models (LLMs) is crucial for ensuring reliability, particularly in complex, multi-step interactions between humans and LLMs. Traditional self-consistency methods often miss subtle semantic changes in natural language and functional shifts in code or equations, which can accumulate over multiple transformations. To address this, we propose \textit{ConsistencyChecker}, a tree-based evaluation framework designed to measure consistency through sequences of reversible transformations, including machine translation tasks and AI-assisted programming tasks. 
In our proposed framework, nodes represent distinct text states, while edges correspond to pairs of inverse operations. 
Dynamic and LLM-generated benchmarks ensure a fair assessment of the model’s generalization ability and eliminate benchmark leakage. Consistency is quantified based on similarity across different depths of the transformation tree. Experiments on eight models from various families and sizes show that ConsistencyChecker can distinguish the performance of different models. Notably, our consistency scores, computed entirely without using WMT paired data, correlate strongly ($r$ > 0.7) with WMT 2024 auto-ranking, demonstrating the validity of our benchmark-free approach. Our implementation is available at~\url{https://github.com/ulab-uiuc/consistencychecker}.
\vspace{-3mm}
\end{abstract}

\begin{figure}[t!]
    \centering
    \includegraphics[width=0.4\textwidth]{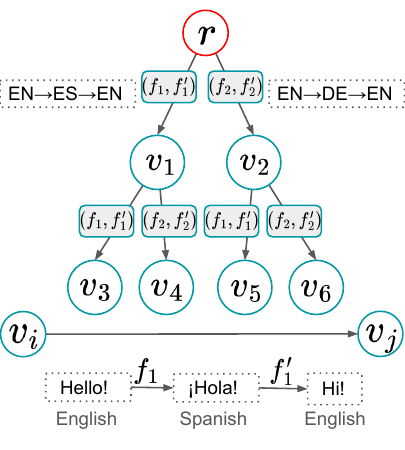}
    \vspace{-3mm}
    \caption{\textbf{Overview of the ConsistencyChecker.} It shows a self-consistency tree for evaluating LLMs on machine translations. The root node ($r$) is the initial English sentence generated by the \textit{evaluator}. Subsequent nodes ($v_i$) are produced by the \textit{evaluatee} using transformation pairs, such as English$\rightarrow$Spanish$\rightarrow$English ($f_1, f_1'$) and English$\rightarrow$German$\rightarrow$English ($f_2, f_2'$). The evaluation framework supports multilingual translations (\textit{e.g.}, French, Czech, Japanese) and can be extended to programming tasks.}
    \label{fig:fig1}
    \vspace{-4mm}
\end{figure}

\section{Introduction}

Large Language Models (LLMs) have emerged as transformative tools in artificial intelligence, demonstrating remarkable capabilities in natural language understanding \citep{minaee2024}, generation \citep{minaee2024}, and complex reasoning \citep{brown2020,chowdhery2022}. These models are increasingly deployed in applications such as multilingual translation pipelines~\citep{xu2023paradigm} and automated code generation~\citep{jiang2024survey}, all of which require maintaining semantic and functional consistency across multiple interactions. A fundamental research question thus arises: \textit{Can we effectively evaluate the consistency of LLMs over multi-step transformations}?

\vspace{1mm}
\xhdr{Challenges for evaluating consistency}
Evaluating LLM consistency across multi-step transformations is inherently difficult. Existing methods, such as self-consistency sampling~\citep{min2023beyond,wang2022self}, which typically assess agreement between outputs for a single prompt, could fail to capture how meaning may drift over a sequence of transformations. This limitation is critical in tasks like machine translation and code generation, where minor changes at each iteration, though locally coherent, can accumulate into major inconsistencies after multiple steps. For example, a translated sentence may gradually shift semantically, and a code snippet may become non-executable after multiple transformations. Effective evaluation must therefore assess \textit{semantic} and \textit{functional} consistency across the multi-step trajectory. For simplicity, we consider semantic consistency to be a subset of functional consistency, which mostly applies to natural language, whose functional consistency is equivalent to its semantic information-keeping consistency.

\vspace{1mm}
\xhdr{Challenges in evaluating generalization abilities} Most benchmarks aim to evaluate how LLMs perform on unseen data in specific domains, which is to assess their generalization capabilities. Existing evaluation approaches primarily rely on dataset-centric benchmarks (\textit{e.g.}, HumanEval~\citep{chen2021evaluating} for code generation). The dataset construction process includes human annotation or crawling of real-world data from the Internet. However, due to the vast and opaque nature of LLM training data, it has become increasingly difficult to determine whether benchmark examples were truly unseen during training. Several studies have found evidence of benchmark leakage~\cite{xu2024benchmarkingbenchmarkleakagelarge,zhou2025lessleakbenchinvestigationdataleakage}, where evaluation data overlaps with training corpora, undermining the reliability of performance metrics. Therefore, there is a growing need for novel benchmarking methodologies that dynamically generate evaluation data to ensure a more faithful assessment of model generalization.

\vspace{1mm}
\xhdr{Novel tree-based benchmark-free evaluation} To address these limitations, we introduce \textit{ConsistencyChecker}, a \textit{tree-based} and \textit{benchmark-free} evaluation framework that operates without reference datasets, enabling cheap and domain-agnostic assessments of LLM generalization capabilities. At its core, \textit{ConsistencyChecker} constructs self-consistency trees, where each node represents an LLM-generated state in transformations (\textit{e.g.}, output of one translation step), each edge denotes a pair of inverse operations, and each path in the self-consistency tree represents a multi-step transformation sequence operated by the LLM. For instance, starting with an English sentence, we perform a round-trip translation (English-French-English) executed by the LLM, yielding a slightly varied English sentence as a subsequent node. By analyzing consistency across nodes at different depths, \textit{ConsistencyChecker} quantifies an LLM's reliability in retaining consistency (\textit{e.g.} a sentence still conveys the same information, or code still gives the same output given the same inputs) through iterative transformations. \textit{ConsistencyChecker} highlights two key innovations: (1) utilizing a tree structure to comprehensively evaluate consistency across multi-step transformation processes; (2) utilizing dynamically sampled data from LLMs to prevent benchmark leakage during evaluation.

\vspace{1mm}
\xhdr{Key findings}
Based on our experimental results on both machine translation and AI-assisted programming tasks, we find that:(1) \textit{ConsistencyChecker} can effectively distinguish the performance of different models and assign them varying scores, demonstrating its utility as a distinguishable metric for both tasks; (2) GPT-4o-mini ranks highest in machine translation tasks, while Qwen-2.5-32B performs best in code generation tasks, indicating that evaluation results align with these models' performance on other benchmarks; (3) \textit{ConsistencyChecker} can operate without relying on predefined benchmarks and still produces rankings highly correlated with expensive evaluations relying on paired data on the WMT 2024 benchmark.

\section{Related Works}

\vspace{1mm}
\xhdr{Round-trip consistency in translations}
Machine translation evaluation has evolved from early rule-based systems~\citep{weaver1952} to modern transformer-based approaches~\citep{vaswani2017}. Traditional metrics like BLEU~\citep{papineni2002}, ROUGE~\citep{lin2004}, and TER~\citep{snover2006} rely on reference translations, while benchmarks like WMT 2024~\citep{barrault2019} depend on parallel corpora crawled from internet resources, often unavailable in low-resource languages or specialized domains. Round-trip translation, where text is translated between languages and back~\citep{van2006}, provides a benchmark-free quality checker analogous to our transformation chains. ConsistencyChecker generalizes this concept, extending it beyond translation to arbitrary functional-preserving operations (\textit{e.g.}, AI-assisted programming), thereby enabling evaluation in scenarios where parallel data or reference outputs are absent.

\vspace{1mm}
\xhdr{Consistency-based evaluation}
Recent LLM evaluation frameworks, such as Divide-Conquer-Reasoning~\citep{cui2024divide} and MT-Eval~\citep{kwan2024mt}, improve consistency assessment by breaking down tasks and using dynamic benchmarks. However, they often miss tracking semantic or functional drift across complex, multi-step transformations~\citep{wang2024latent}. Our work builds on these ideas by introducing an explicit tree-structured evaluation, which systematically measures consistency at every step and path. Uniquely, we also use LLMs to generate dynamic benchmarks, enabling comprehensive and scalable consistency evaluation beyond existing methods.

\vspace{1mm}
\xhdr{Formal verification method}
The development of reliable systems has traditionally depended on formal verification techniques such as model checking~\citep{clarke1999}, theorem proving~\citep{nipkow2002}, and SMT solvers~\citep{barrett2018}, all of which require significant computational resources and domain-specific specifications. The recent rise of LLMs for code generation has led to benchmarks like CodeXGLUE~\citep{lu2021} and ProblemSolving~\citep{hendrycks2021}, but these approaches rely on curated datasets, which limits their general applicability. Inspired by both formal methods and the input/output-based evaluation used by online programming platforms like LeetCode, we instead use a set of test inputs to verify the functionality of LLM-edited Python code. Unlike traditional benchmarks, our approach does not require predefined correct outputs; instead, we concatenate and compare the outputs to assess functional consistency between different nodes.

\section{Preliminary}
\label{sec:preliminary}
In this section, we first provide detailed information on the definition of the self-consistency tree, which is the key concept for our proposed ConsistencyChecker framework. Furthermore, we introduce two derived concepts of the self-consistency tree: paths and forests. These concepts are necessary for the calculation of our proposed consistency score.
\subsection{Self-consistency Tree}

To evaluate how well LLMs preserve consistency across reversible transformations, we begin by introducing our definition of operation, which refers to real-world tasks like translation or code editing defined by prompts and will be executed by the evaluatee LLM (the LLM to be evaluated). These operations are paired with their inverses to enable round-trip consistency checks. We then define nodes as intermediate model outputs and edges as paired operations. Together, they form the self-consistency tree, a structure that models the multi-step transformation process and allows consistency to be evaluated at different path lengths.

\begin{figure}[t!]
    \centering
    \hspace{-2mm}
    \includegraphics[width=0.5\textwidth]{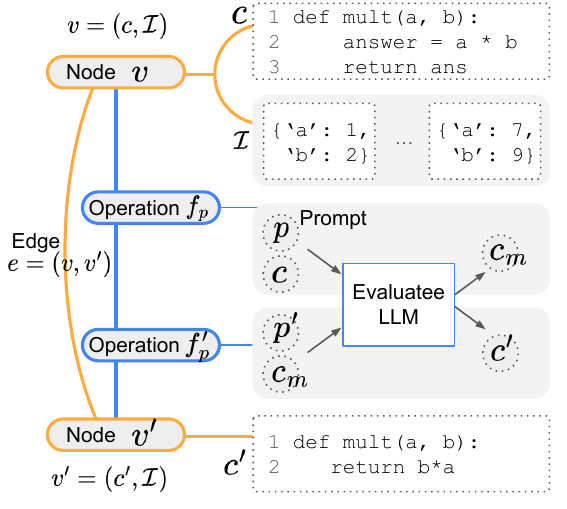}
    \vspace{-6mm}
    \caption{\textbf{Key concepts in the self-consistency tree (operation, node, and edge)}. It provides a concrete example of the self-consistency tree in the AI-assisted programming task. The node $v$ contains a function that returns the product of two positive integers, and $\mathcal{I}$ is its set of inputs. The prompt $p$ asks the evaluatee LLM to rewrite the code in node $v$ to use a looped sum instead of just multiplying, while $p'$ asks the same LLM to alter it back to simply multiplying.}
    \label{fig:fig2}
    \vspace{-4mm}
\end{figure}

\vspace{1mm}
\xhdr{Operation (basic block)}
An operation is defined as a prompt-driven transformation performed by the evaluated LLM. For each prompt $p$, we define an operation $f_p$ and its inverse $f_p'$, such that applying $f_p$ followed by $f_p'$ ideally recovers the original input. Formally,
\begin{equation}
f_p: c \rightarrow c',
\end{equation}
where $c$ is the input (\textit{e.g.}, code or text) and $c'$ is the output after the transformation. The inverse operation $f_p'$ is constructed so that $f_p'(f_p(c)) \approx c$ if the evaluated LLM is ideal.

\vspace{1mm}
\xhdr{Node (LLM-generated state)}
Each node $v = (c, \mathcal{I})$ is a tuple representing a single LLM generation and its associated test inputs:
(1) $c \in \Sigma^*$ is the generated content (\textit{e.g.}, code, function, text);
(2) $\mathcal{I} = [i_1, \ldots, i_n]$ is a list of test inputs used to verify functional behavior.
Here, $\Sigma^*$ denotes the set of all possible strings. Regardless of output format (natural language, code, etc.), we treat each generated content $c$ as an executable function for the evaluation of unified consistency. For semantic similarity evaluation in machine translation tasks, we consider it as a special case where $I$ is an empty set and the executed outcomes remain $c$.

\vspace{1mm}
\xhdr{Edge (paired operations)}
Each edge represents an operation and its inverse, $(f_p, f_{p'})$. It connects two nodes that are evaluated on the same set of test inputs, ensuring that functional consistency can be assessed under identical conditions. Formally, an edge $e_{ij} = (v_i, v_j)$ connects nodes $v_i = (c_i, \mathcal{I}_i)$ and $v_j = (c_j, \mathcal{I}_j)$ if and only if
\begin{equation}
\vspace{-1.5mm}
\left\{
\begin{aligned}
    \hspace{6pt}
    \mathcal{I}_i &= \mathcal{I}_j, \hspace{6pt}
     \\
    c_j &= f_{p'}(f_p(c_i)).
\end{aligned}
\right.
\end{equation}

\vspace{1mm}
\xhdr{Self-consistency tree}
Combining these concepts, a self-consistency tree $\mathcal{T} = (\mathcal{V}, \mathcal{E})$ models the multi-step process by which an LLM transforms a root input through a sequence of operations. This hierarchical structure enables comprehensive measurement of how well the LLM preserves semantics or functionality over multiple transformation steps, with verification performed using the shared input set $\mathcal{I}$. Figure~\ref{fig:fig1} illustrates the construction of nodes, edges, and operations. The tree generation process, detailed in Algorithm~\ref{alg:tree_generation}, uses operation pairs and root nodes. Notably, the branching factor at each layer of the tree is determined by the number of available operation pairs.

\subsection{Derived Concepts: Path and Forest}
Based on the definition of the self-consistency tree, we can extend it and define more useful concepts on top of it. These derived concepts help us calculate our proposed consistency score.

\vspace{1mm}
\xhdr{Path}
A path in the self-consistency tree is defined as a sequence of nodes $P = (v_1, ..., v_n)$, where each node is sequentially connected with the next through a valid edge in the self-consistency tree. The path represents a series of inverse operations applied by the evaluatee LLM, and its length corresponds to the number of operation pairs (the depth difference between the starting and ending nodes), allowing us to measure how consistency degrades across a certain number of transformation steps.

\vspace{1mm}
\xhdr{Forest}
A forest $\mathcal{F} = \{\mathcal{T}_1, ..., \mathcal{T}_M\}$ consists of multiple self-consistent trees, each rooted at a different initial node. These initial nodes are generated using distinct meta-prompts - \textit{e.g.}, ``Write a 400-word paragraph in English about recent advances in computer science''. This design allows us to analyze LLM consistency across diverse tasks and domains by comparing patterns across trees in the forest.

\vspace{-2mm}
\section{ConsistencyChecker: Tree-based LLM evaluation for Generalization Ability}
Based on the definitions in Section~\S\ref{sec:preliminary}, we now detail the construction of consistency scores in ConsistencyChecker. We begin by defining the similarity score for node pairs, then extend this to path-level, tree-level, and finally forest-level scores, with each step building upon the previous one.

\vspace{1mm}
\xhdr{Node-pair similarity score}
To quantify the functional and semantic similarity between two connected nodes $v_i = (c_i, \mathcal{I})$ and $v_j = (c_j, \mathcal{I})$ that share the same test inputs $\mathcal{I}$, we first obtain their execution outputs: $o_i = \text{exec}(c_i, \mathcal{I})$ and $o_j = \text{exec}(c_j, \mathcal{I})$. The similarity score $\text{sim}(v_i, v_j)$ is then computed between $o_i$ and $o_j$, using measures such as the cosine similarity of semantic embeddings or the BLEU score.

\begin{algorithm}[t!]
\caption{Tree Generation Algorithm}
\label{alg:tree_generation}
\begin{algorithmic}[1]
\Require Root node $r$, max depth $D$, operation pairs $((p_1, p_1'), \cdots, (p_k, p_k'))$
\Ensure Tree $\mathcal{T} = (\mathcal{V}, \mathcal{E})$
\State $\mathcal{V} \gets \{\, r \,\}$
\State $\mathcal{E} \gets \emptyset$
\For{$d \gets 1$ to $D$}
    \State $U \gets \{\, v \in \mathcal{V} \mid \mathcal{D}(v) = d-1 \,\}$
    \For{each $v \in U$}
        \For{$i \gets 1$ to $k$}
            \State $c' \gets f_{p_i'}(f_{p_i}(v.c))$
            \State $v' \gets (c', v.\mathcal{I})$
            \State $\mathcal{V} \gets \mathcal{V} \cup \{\, v' \,\}$
            \State $\mathcal{E} \gets \mathcal{E} \cup \{\, (v, v') \,\}$
        \EndFor
    \EndFor
\EndFor
\State \Return $(\mathcal{V}, \mathcal{E})$
\end{algorithmic}
\end{algorithm}

\vspace{1mm}
\xhdr{Path-level consistency score} To evaluate whether a node still retains core functionalities after iterative transformations, we measure the end-to-end consistency between the initial and final nodes in a path. For a path on the tree \( P = (v_1, \cdots, v_n) \), the consistency score defined for that is:
\begin{equation}
C(P) \coloneqq \text{sim}(v_1, v_n)
\end{equation}
This metric evaluates whether the final output $v_n$ remains functionally equivalent to the original $v_1$ after a sequence of transformations. For example, for code generation, applying operations of "add logging functionality" followed by "remove logging functionality" should leave the core sorting functionality unchanged, as reflected by this score.

\vspace{1mm}
\xhdr{Tree-level consistency score}
To assess how well an LLM preserves functional consistency over fixed-length transformation sequences, we define a tree-level consistency score with the help of the path-level consistency score. For a given depth $n$, we consider all valid $n$-step paths in the self-consistency tree as a set: $\mathcal{P}_n(\mathcal{T}) = \left\{ (v_1, \ldots, v_n) ~\middle|~ (v_i, v_{i+1}) \in \mathcal{E} \text{ for } i = 1,\ldots,n \right\}$. The tree-level consistency score is the average of path-level consistency scores over all such paths:
\begin{equation}
C_n(\mathcal{T}) \coloneqq \frac{1}{|\mathcal{P}_n(\mathcal{T})|} \sum_{P \in \mathcal{P}_n(\mathcal{T})} C(P)
\end{equation}
where $C(P)$ refers to the consistency score at the path level. This metric captures the model’s ability to maintain core functionality across all $n$-step transformation sequences, reflecting both cumulative error and path-specific divergence.

\vspace{1mm}
\xhdr{Forest-level consistency score}
To assess robustness and allow the measurement of generalization ability at specific domains, we define a consistency score over a forest of self-consistency trees $\mathcal{F} = \{\mathcal{T}_1, \ldots, \mathcal{T}_M\}$. The score is computed as:
\begin{equation}
C_n(\mathcal{F}) \coloneqq \frac{1}{M} \sum_{m=1}^{M} C_n(\mathcal{T}_m),
\end{equation}
where $C_n(\mathcal{T})$ denotes the consistency at path length $n$ of tree $\mathcal{T}$. Each tree is rooted at a node generated by the evaluator model. 
A forest with a high $C_1(\mathcal{F})$ but a low $C_3(\mathcal{F})$ indicates consistent performance for short transformation chains but degradation in longer transformation sequences. 

\vspace{1mm}
\xhdr{ConsistencyChecker}
In practice, setting a large value for $n$ increases computational cost, while a small $n$ leads to inaccurate estimates that overlook accumulated errors in multi-step transformations. To balance these trade-offs, we select $n=3$ and use the forest-level consistency score $C_3(\mathcal{F})$ as the final metric in our \textit{ConsistencyChecker} framework.

\section{Evaluation Tasks and Meta Prompts}

\begin{table}[t]
\centering
\small
\caption{\textbf{Meta prompts for evaluation tasks}. Instructions about YAML and formatting are abbreviated.}
\vspace{-2mm}
\begin{tabular}{p{0.95\linewidth}}
\toprule
\textbf{Machine translation} \\
\midrule
Write a 400 word, complicated English paragraph that might appear on a news website. \\
Please do this in a function way, e.g. provide a function called "main" that returns the content as a string. \\
\midrule
\textbf{AI-assisted programming} \\
\midrule
Write a LeetCode-Hard style problem. The problem must be super hard, even a graduate student in computer science will struggle to solve it. \\
Do not attempt to generate long, nested dicts. But it will require a very long and complicated solution. \\
The execution time should be very short. However, it does not need to be super long. It can be shorter, but it must be really hard. \\
Please do this in a functional way, \textit{e.g.}, provide a function called "main" that returns the intended answer. \\
\bottomrule
\label{meta-prompts}
\end{tabular}
\vspace{-9mm}
\end{table}
We evaluate ConsistencyChecker on two tasks: \textit{machine translation} and \textit{AI-assisted programming}. Each task is specified by a meta-prompt that generates a set of root nodes and paired operations for constructing self-consistency trees. In these cases, nodes correspond to either natural language sentences or executable code snippets. These tasks are representative, as they cover the full spectrum of node types—from static text with no external dependencies to code that serves as a solution to a LeetCode-style problem. The machine translation task focuses on assessing semantic consistency, while the AI-assisted programming task focuses on evaluating functional consistency. We set a time limit of 2 seconds per case in the test inputs to ensure that the verification process can end in a predictable time.

\vspace{1mm}
\xhdr{Machine translation} The root node of a translation task is a paragraph randomly sampled. Operations involve translating this paragraph into and from another language. We sample three languages from an evaluator, which are typically French, Spanish, and German. This setup ensures that the task captures the nuances of multilingual translation across diverse linguistic contexts.

\vspace{1mm}
\xhdr{AI-assisted programming} The root node for each programming task consists of a LeetCode-style coding problem and 20 associated test inputs. Operations are defined as transformations that prompt the model to implement the solution in a different, but functionally equivalent, way. This setup enables us to evaluate the model’s ability to generate code that remains functionally and semantically consistent across varying levels of abstraction and complexity.

\vspace{1mm}
\xhdr{Connections between two tasks.} Although the translation and coding tasks differ in complexity and domain, both are implemented within the same framework. For translation, the function takes no inputs and simply returns the target natural language text. For the coding task, the function accepts inputs as required by the problem. This unified design makes our framework easily extensible to other tasks, such as step-by-step arithmetic, which can also be represented as Python code.

\section{Experimental Settings}
\vspace{1mm}
\xhdr{Evaluator models} Evaluator models determine the quality of dynamic benchmark generation. All experiments, except those in Table~\ref{combined-table-translation-and-code-l3}, use Qwen-2.5-72B as the evaluator model to generate dynamic benchmarks. For both tasks, we use the same evaluator-generated benchmarks across all models to ensure fair comparison and minimize the variability that fully on-the-fly benchmarks could introduce. Specifically, the evaluator model generates 10 root nodes for each task, which are then used consistently to evaluate all LLMs.

\vspace{1mm}
\xhdr{Evaluatee models} 
Using LLM-generated benchmarks, we evaluate a diverse set of language models across different families and sizes, including GPT-4o-mini~\cite{openai2024gpt4ocard}, Qwen-2.5 (1.5B, 7B, 14B, 32B, 72B), and LLaMA-3.1 (8B, 70B)~\cite{grattafiori2024llama}, on both machine translation and AI-assisted programming tasks. Additionally, we assess the correlation between ConsistencyChecker and WMT 2024 metrics for machine translation benchmarks on specific language pairs. The models evaluated in this comparison include Claude-3.5-Sonnet~\cite{TheC3}, GPT-4, Gemini-1.5-Pro~\cite{geminiteam2024gemini15unlockingmultimodal}, Mistral-Large~\cite{jiang2023mistral7b,jiang2024mixtralexperts}, LLaMA-3-70B, and Phi-3-Medium~\cite{abdin2024phi3technicalreporthighly}.

\vspace{1mm}
\xhdr{ConsistencyChecker settings}
For the calculation of ConsistencyChecker, we fix $n = 3$, $M=10$, and use embedding-based metrics such as NV-Embed-v2\footnote{\url{https://huggingface.co/nvidia/NV-Embed-v2}} to compute similarity scores between node pairs. We then calculate the forest-level consistency score as our main result. All reported scores are presented as means with standard deviations to reflect variability across evaluation runs.

\begin{table}[t]
\centering
\small
\caption{\textbf{Evaluation results with ConsistencyChecker.} Each score represents LLMs’ ability to maintain semantic similarity in machine translation tasks and preserve functional consistency in programming tasks. Higher score indicates larger ratios for preserving the original information after transformations.}
\vspace{-2mm}
\scalebox{1.0}{
\begin{tabular}{lcc}
\toprule
\shortstack{\textbf{Model}} & 
\shortstack{\textbf{Translation}$\uparrow$} & \shortstack{\textbf{Programming}$\uparrow$} \\
\midrule
GPT-4o-mini       & \textbf{98.0$_{\pm0.0}$} & 76.5$_{\pm2.7}$ \\
\midrule
Qwen-2.5-1.5B     & 80.3$_{\pm0.5}$         & 63.4$_{\pm1.4}$ \\
Qwen-2.5-7B       & 90.0$_{\pm0.8}$         & 71.7$_{\pm0.4}$ \\
Qwen-2.5-14B      & 94.7$_{\pm0.1}$         & 79.9$_{\pm1.0}$ \\
Qwen-2.5-32B      & 96.4$_{\pm0.0}$         & \textbf{85.1$_{\pm1.1}$} \\
Qwen-2.5-72B      & 97.2$_{\pm0.0}$         & 77.0$_{\pm1.9}$ \\
\midrule
LLaMA-3.1-8B      & 67.5$_{\pm3.0}$         & 60.4$_{\pm1.0}$ \\
LLaMA-3.1-70B     & 71.9$_{\pm3.2}$         & 83.5$_{\pm1.0}$ \\
\bottomrule
\label{merged-table}
\end{tabular}
}
\vspace{-5mm}
\end{table}

\setlength{\tabcolsep}{3pt}
\begin{table}[htp]
    \caption{\textbf{Comparison between ConsistencyChecker and other metrics in WMT 2024 benchmarks.} ConsistencyChecker has a high correlation with reliable translation metrics like CometKiwi and AutoRank without utilizing paired data in WMT 2024.}
    \vspace{-2mm}
    \label{consistency-vs-wmt-translation-1}
    \small 
    \centering
    \scalebox{1.0}{
    \begin{tabular}{lccc}
    \toprule
        \textbf{Model} 
        & \textbf{Ours $\uparrow$} 
        & \textbf{CometKiwi $\uparrow$} 
        & \textbf{AutoRank $\downarrow$} \\
    \midrule
    \multicolumn{4}{c}{\textbf{Czech–Ukrainian}} \\
    \midrule
    Claude-3.5-Sonnet & 98.1 & 68.3 & 1.7 \\
    GPT-4             & 96.4 & 67.7 & 2.0 \\
    Gemini-1.5-Pro    & 97.5 & 66.8 & 2.0 \\
    Mistral-Large     & 96.5 & 66.6 & 2.3 \\
    LLaMA-3-70B        & 95.9 & 66.1 & 2.6 \\
    Phi-3-Medium      & 44.9 & 42.5 & 9.1 \\
    \midrule
    \multicolumn{4}{c}{\textbf{English–Chinese}} \\
    \midrule
    Claude-3.5-Sonnet & 99.3 & 70.3 & 1.7 \\
    Gemini-1.5-Pro    & 99.3 & 69.8 & 1.8 \\
    GPT-4             & 98.1 & 69.3 & 2.0 \\
    Mistral-Large     & 97.5 & 66.5 & 2.8 \\
    LLaMA-3-70B         & 97.9 & 66.2 & 2.8 \\
    Phi-3-Medium      & 96.6 & 64.8 & 3.1 \\
    \bottomrule
    \end{tabular}
    }
    \vspace{-6mm}
\end{table}

\begin{figure*}[t]
    \centering
    \begin{subfigure}[b]{0.5\linewidth}
        \centering
        \caption{Machine translation consistency scores (\%).}
        \includegraphics[width=0.9\linewidth]{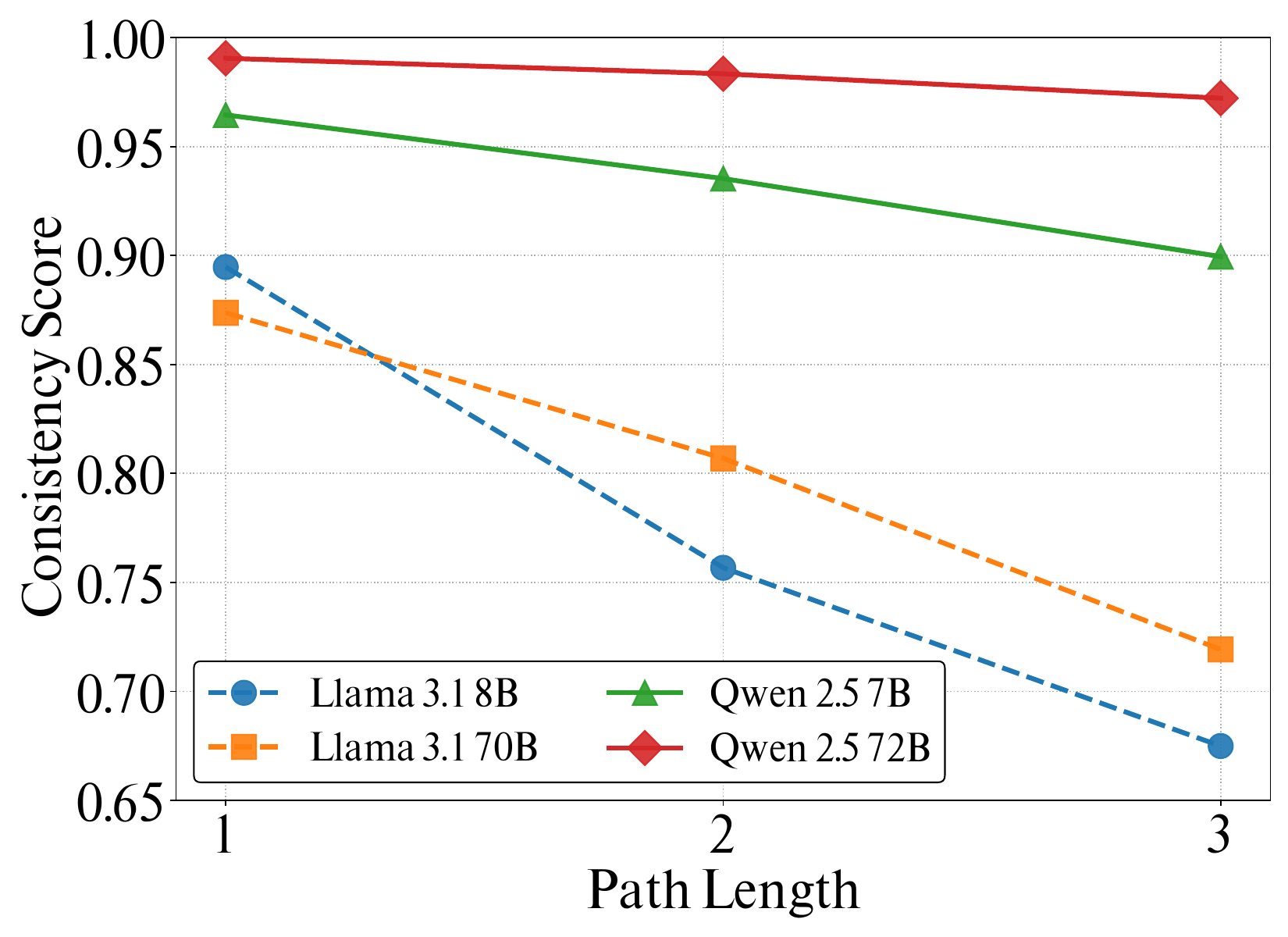}
        \label{fig:translate-scores}
    \end{subfigure}\hfill
    \begin{subfigure}[b]{0.5\linewidth}
        \centering
        \caption{AI-assisted programming consistency scores (\%).}
        \includegraphics[width=0.9\linewidth]{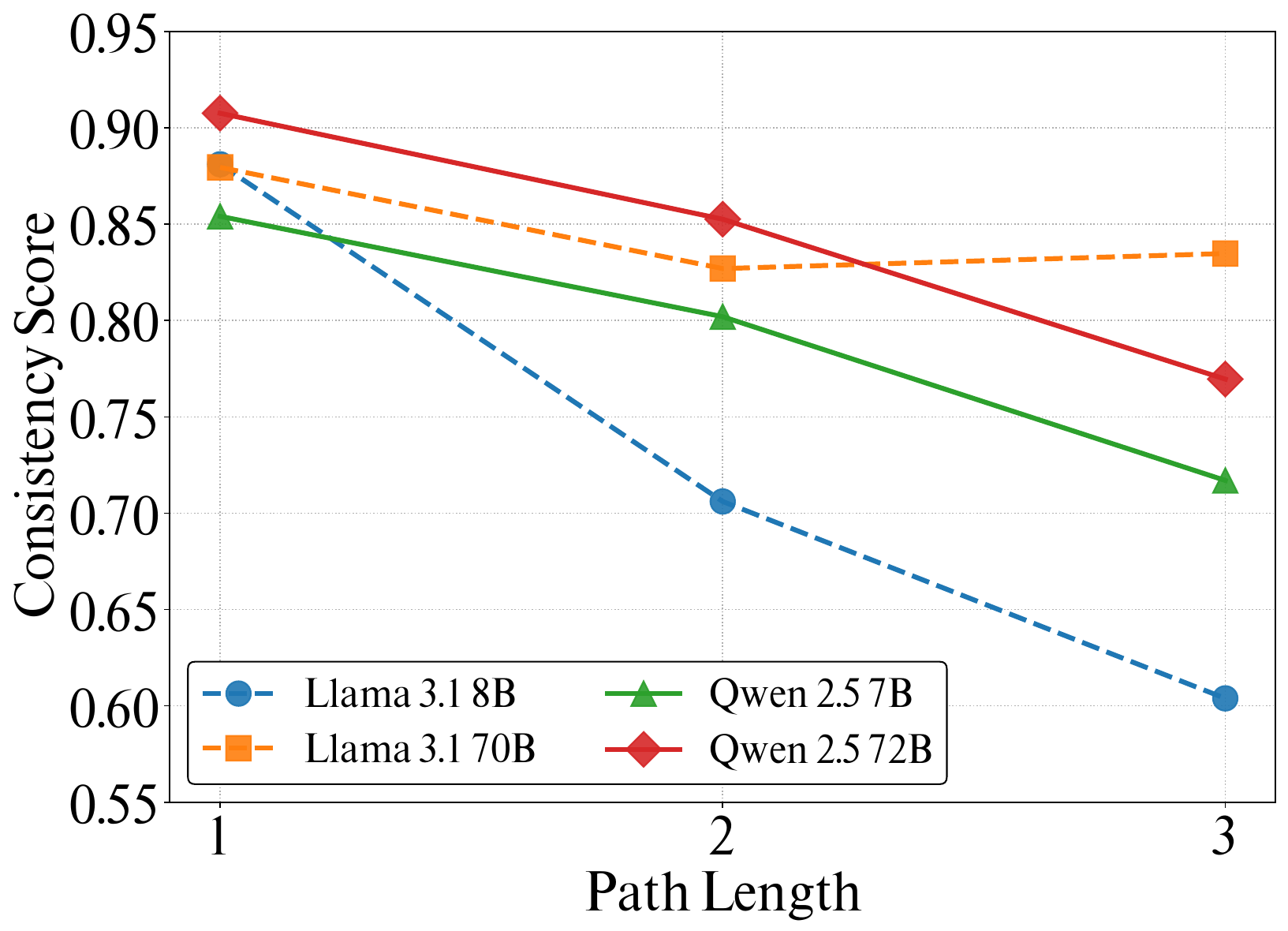}
        \label{fig:code-scores}
    \end{subfigure}
    \vspace{-7mm}
    \caption{\textbf{Ablation study on path lengths.} For most models and for both translation and programming tasks, the consistency score becomes smaller when the path becomes longer.}
    \vspace{-2mm}
    \label{fig:layer_number}
\end{figure*}

\section{Experimental Results}
We show experimental results of 8 evaluated models on both machine translation and AI-assisted programming tasks, and analyze the trends and insights for the two tasks separately.

\vspace{1mm}
\xhdr{ConsistencyChecker is reliable for evaluating machine translation tasks} As shown in Table~\ref{merged-table}, larger models consistently achieve higher consistency scores than smaller models within the same family for machine translation tasks. For example, Qwen-2.5-72B achieves a score of 97.2, which is 21.1\% higher than Qwen-2.5-1.5B. Similarly, LLaMA-3.1-70B outperforms LLaMA-3.1-8B by 6.5\%. Overall, models from the Qwen family generally achieve higher consistency scores than those from the LLaMA family, indicating that Qwen models produce more consistent results. Specially, we find GPT-4o-mini achieves the highest consistency score of 98.0, indicating the highest quality of it on machine translation tasks.

\vspace{1mm}
\xhdr{ConsistencyChecker is reliable for evaluating AI-assisted programming tasks}
In AI-assisted programming tasks, larger models generally achieve higher functional consistency than smaller ones within the same model family, with Qwen-2.5-72B as an exception. Concretely, Qwen-2.5-32B scores 34.2\% higher than Qwen-2.5-1.5B, while LLaMA-3.1-70B outperforms LLaMA-3.1-8B by 38.2\%. These results underscore the importance of model scale for robust performance on complex transformation tasks.

\vspace{1mm}
\xhdr{ConsistencyChecker reaches high correlation with metrics in WMT 2024 without using its paired data} We compare ConsistencyChecker with the WMT 2024~\citep{kocmi2024preliminary} automatic rankings, focusing on two widely used metrics: CometKiwi (we refer to CometKiwi-DA-XL~\citep{rei2023scalingcometkiwiunbabelist2023}) and AutoRank (the average of normalized CometKiwi and MetricX~\citep{juraska2024metricx} scores). Both metrics are computationally expensive, requiring large-scale pairwise human data or extensive model inference. In contrast, our approach is highly cost-efficient, as it does not require parallel data or data crawling—benchmarks are dynamically generated by LLMs. As shown in Table~\ref{consistency-vs-wmt-translation-1}, we evaluate six representative LLMs—Claude-3.5-Sonnet, Gemini-1.5-Pro, GPT-4, LLaMA-3-70B, Mistral-Large, and Phi-3-Medium. For both CometKiwi and AutoRank, at least four out of six models are ranked similarly under ConsistencyChecker and WMT 2024. Pearson’s correlation coefficients, which are above 0.8 for both language pairs, provide strong evidence of the close agreement between ConsistencyChecker and established evaluation metrics. Detailed results are provided in Appendix~\S\ref{appendix:correlation-study}.

\section{Ablation Study}
The path length $n$ that we utilize for the calculation of ConsistencyChecker and the evaluator model that we rely on for dynamic benchmark generation are two key design elements that affect the performance of ConsistencyChecker. Therefore, we conduct ablation studies on them.

\vspace{1mm}
\xhdr{Ablation on path length}
Figure~\ref{fig:layer_number} shows that consistency scores generally decline as path length increases, reflecting greater transformation complexity. For example, in the machine translation task, the LLaMA-3.1-8B model shows a 24.6\% drop in consistency from path length 1 to 3, while the Qwen-2.5-72B model shows a smaller 1.9\% decrease. This trend holds across most models, regardless of family or size, though the rate of decline varies. In the AI-assisted programming task, the Qwen-2.5-7B model exhibits a 16.0\% reduction, again confirming that errors tend to accumulate over longer transformation paths. Smaller drops in the consistency score indicate stronger self-consistency, as the model introduces fewer errors per operation. An exception occurs with the LLaMA-3.1-70B model in the programming task, where the consistency score is higher at path length 3 than at 2. This may be due to the model detecting and correcting syntax errors or bugs that appeared at intermediate steps.

\vspace{1mm}
\xhdr{Ablation on evaluator model}
As shown in Table~\ref{combined-table-translation-and-code-l3}, different evaluator models (Qwen-2.5-7B and Qwen-2.5-72B) produce similar rankings across evaluated LLMs. For instance, in the translation task at path length 3, Qwen-2.5-7B assigns a score of 76.8, while Qwen-2.5-72B gives a slightly higher score of 78.2. In the programming task, for example, Qwen-2.5-7B evaluated by itself scores 72.2, while evaluation by Qwen-2.5-72B yields 71.7 (0.7\% difference). We find that such an alignment becomes more pronounced at longer path lengths, where consistency patterns are clearer. Although absolute consistency scores may vary between evaluators, relative trends remain stable. 

\section{Discussion}

In our main experiments, we use embedding‐based models (NV-embed-v2) to compute node-pair similarity $\text{sim}(\cdot,\cdot)$ and calculate forest-level consistency score $C_3(\mathcal{F})$ based on that. However, BLEU~\citep{papineni2002} remains a classical and widely adopted measure of translation quality. Therefore, we discuss their differences here.

\vspace{1mm}
\xhdr{BLEU is highly correlated but slightly worse than embedding-based models for machine translations} When evaluating translation across the five language pairs defined in WMT 2024, the BLEU-based consistency score is well correlated with WMT benchmark metrics, achieving Pearson correlations above 0.7 with three reference metrics in four out of five pairs. Furthermore, the BLEU-based $C_3(\mathcal{F})$ consistency score is also strongly correlated with the embedding-based $C_3(\mathcal{F})$, with average Pearson and Spearman correlations of 0.89 and 0.87, respectively, across all five pairs. However, embedding-based consistency scores are slightly more correlated with WMT benchmark metrics than BLEU-based scores when averaged across the five language pairs. Full evaluation results are provided in Appendix~\S\ref{appendix:correlation-study}.

\vspace{1mm}
\xhdr{BLEU is surprisingly useful for evaluating programming tasks} For the AI-assisted programming task, the correlations were notably strong: Pearson coefficients reached 0.98 ($C_1(\mathcal{F})$), 0.97 ($C_2(\mathcal{F})$), and 0.99 ($C_3(\mathcal{F})$). This high correlation is unexpected, as BLEU is not designed to measure output functionality but is designed to measure n-gram overlaps. However, these results indicate that in this context, BLEU captures meaningful distinctions between outputs, likely because nodes are either fully functional (producing identical outputs for the same inputs), entirely different, or unexecutable due to code errors. As a result, even when outputs are not natural language, BLEU can effectively differentiate between them.

\begin{table}[t]
  \caption{\textbf{Ablation study on evaluator models.} (Top) machine translation consistency scores (\%) at complexity level $C_3(\mathcal{F})$ for model scales (7B–72B). (Bottom) AI-assisted programming consistency scores (\%) at complexity level $C_3(\mathcal{F})$ for code transformations.}
  \vspace{-2mm}
  \label{combined-table-translation-and-code-l3}
  \begin{center}
    \begin{small}
      \setlength{\tabcolsep}{6pt}
      \begin{tabular}{l|cc}
        \toprule
        \diagbox{Evaluatee}{Evaluator}
          & Qwen-2.5-7B
          & Qwen-2.5-72B \\
        \midrule
        \multicolumn{3}{c}{\textbf{Machine translation}} \\
        \midrule
        Qwen-2.5-7B     & 76.8$_{\pm5.1}$  & 90.0$_{\pm0.8}$ \\
        Qwen-2.5-72B    & \textbf{78.2}$_{\pm7.3}$  & \textbf{97.2}$_{\pm0.0}$ \\
        LLaMA-3.1-8B    & 61.7$_{\pm4.8}$  & 67.5$_{\pm3.0}$ \\
        LLaMA-3.1-70B   & 30.8$_{\pm9.2}$  & 71.9$_{\pm3.2}$ \\
        \midrule
        \multicolumn{3}{c}{\textbf{AI-assisted programming}} \\
        \midrule
        Qwen-2.5-7B     & 72.2$_{\pm1.9}$  & 71.7$_{\pm0.4}$ \\
        Qwen-2.5-72B    & 79.4$_{\pm2.3}$  & 77.0$_{\pm1.9}$ \\
        LLaMA-3.1-8B    & 61.0$_{\pm1.1}$  & 60.4$_{\pm1.0}$ \\
        LLaMA-3.1-70B   & \textbf{81.5}$_{\pm2.1}$  & \textbf{83.5}$_{\pm1.0}$ \\
        \bottomrule
      \end{tabular}
      \vspace{-4mm}
    \end{small}
  \end{center}
\end{table}

\begin{table}[!t]
  \caption{\textbf{Consistency score results when using BLEU to build ConsistencyChecker.} We select Qwen-2.5-72B as the evaluator and select different models as evaluatees. We utilize BLEU to build consistency scores.}
  \vspace{-2mm}
  \label{sample-table-merged-bleu-evaluation}
  \begin{center}
    \begin{small}
      \setlength{\tabcolsep}{8pt}
      \begin{tabular}{lccc}
        \toprule
        Evaluatee & \textbf{$C_1(\mathcal{F})$} & \textbf{$C_2(\mathcal{F})$} & \textbf{$C_3(\mathcal{F})$} \\
        \midrule
        \multicolumn{4}{c}{\textbf{Machine translation}} \\
        \midrule
        GPT-4o-mini       & \textbf{86.0}$_{\pm0.0}$ & \textbf{78.6}$_{\pm0.1}$ & \textbf{68.0}$_{\pm0.2}$ \\
        \midrule
        Qwen-2.5-1.5B     & 53.4$_{\pm0.3}$         & 37.6$_{\pm0.3}$         & 25.8$_{\pm0.3}$         \\
        Qwen-2.5-7B       & 70.5$_{\pm0.3}$         & 56.0$_{\pm0.7}$         & 41.9$_{\pm0.7}$         \\
        Qwen-2.5-14B      & 76.3$_{\pm0.1}$         & 65.1$_{\pm0.2}$         & 53.8$_{\pm0.2}$         \\
        Qwen-2.5-32B      & 78.4$_{\pm0.1}$         & 68.2$_{\pm0.2}$         & 57.3$_{\pm0.3}$         \\
        Qwen-2.5-72B      & 81.7$_{\pm0.1}$         & 71.2$_{\pm0.1}$         & 57.4$_{\pm0.2}$         \\
        \midrule
        LLaMA-3.1-8B      & 59.5$_{\pm1.2}$         & 47.6$_{\pm1.1}$         & 36.6$_{\pm0.9}$         \\
        LLaMA-3.1-70B     & 63.1$_{\pm2.0}$         & 57.3$_{\pm1.8}$         & 46.9$_{\pm2.3}$         \\
        \midrule
        \multicolumn{4}{c}{\textbf{AI-assisted programming}} \\
        \midrule
        GPT-4o-mini       & 78.4$_{\pm1.1}$         & 65.0$_{\pm2.5}$         & 48.3$_{\pm8.6}$         \\
        \midrule
        Qwen-2.5-1.5B     & 79.1$_{\pm0.6}$         & 58.0$_{\pm1.3}$         & 23.3$_{\pm3.7}$         \\
        Qwen-2.5-7B       & 64.5$_{\pm0.7}$         & 50.3$_{\pm1.4}$         & 31.4$_{\pm2.4}$         \\
        Qwen-2.5-14B      & 76.8$_{\pm0.6}$         & 60.4$_{\pm2.6}$         & 51.6$_{\pm3.8}$         \\
        Qwen-2.5-32B      & \textbf{79.2}$_{\pm1.1}$ & \textbf{71.9}$_{\pm2.9}$ & \textbf{64.3}$_{\pm4.3}$ \\
        Qwen-2.5-72B      & 78.8$_{\pm1.1}$         & 66.5$_{\pm2.5}$         & 48.9$_{\pm6.4}$         \\
        \midrule
        LLaMA-3.1-8B      & 69.4$_{\pm1.1}$         & 42.9$_{\pm2.4}$         & 11.7$_{\pm2.1}$         \\
        LLaMA-3.1-70B     & 70.5$_{\pm1.2}$         & 58.3$_{\pm1.9}$         & 61.2$_{\pm3.0}$         \\
        \bottomrule
      \end{tabular}
    \end{small}
  \end{center}
  \vspace{-5mm}
\end{table}

\section{Case Study}
In this section, we present two case studies involving different trees with the same root node but with different models for tree generation. 

\vspace{1mm}
\xhdr{Leaf node from different models} The root node contains a function that returns an analysis of AI integration in healthcare systems. We construct self-consistency trees by applying a series of translation operations to the return value of the function: translating it into French, German, and Spanish, then back to English. These trees are evaluated with a path depth of 3, and we compare the outputs of LLaMA-3.1-8B and LLaMA-3.1-70B. The corresponding code for these leaf nodes is shown in Appendix~\S\ref{example-case-study}. Based on the generated content of the leaf nodes, LLaMA-3.1-70B introduces fewer changes (Levenshtein distance: 66) compared to LLaMA-3.1-8B (Levenshtein distance: 215), indicating stronger consistency on larger models. This aligns with their consistency scores provided by ConsistencyChecker, highlighting the robustness of larger models in maintaining meaning through multi-step transformations.

\begin{figure}[t]
    \centering
    \includegraphics[width=0.5\textwidth]{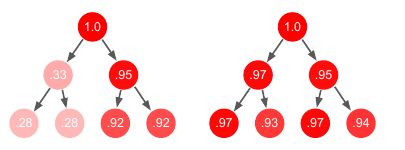}
    \vspace{-6mm}
    \caption{\textbf{Comparison of two self-consistency trees with different performances}. This diagram illustrates how Qwen-2.5-7B (left) and Qwen-2.5-72B (right) performed under the same root node and inverse operation pairs, in machine translation tasks. The diagram shows a simplified case where there are only 2 pairs of inverse operations. We use a lighter shade of red to indicate a smaller embedding similarity with the root node. }
    \label{fig:faildiag}
\end{figure}

\vspace{1mm}
\xhdr{Features of under-performing self-consistency trees} We compare self-consistency trees generated by Qwen-2.5-7B and Qwen-2.5-72B from the same root node to analyze patterns of under-performance from the tree structure. The root consists of an around-290-token English paragraph discussing the impact of European Central Bank policies on global markets. Each model is prompted to translate the text to and from French, German, and Spanish. In the case of Qwen-2.5-7B, the French branch yields a significantly degraded output after round-trip translation, with only a 0.33 embedding similarity score to the root. The resulting English is just around 40 tokens and retains only partial information (see Figure~\ref{fig:faildiag}). Subsequent nodes in this branch show equally low or worse similarity to the root node. By contrast, the German and Spanish branches maintain embedding similarities above 0.86. For Qwen-2.5-72B, all branches - including all child nodes - consistently achieve embedding similarities above 0.90. This illustrates a failure mode of less self-consistent models like Qwen-2.5-7B, and demonstrates how ConsistencyChecker effectively identifies and quantifies such inconsistencies with the help of the tree structure.

\section{Conclusion}
In this work, we propose ConsistencyChecker, an evaluation framework for assessing the generalization ability of LLMs effectively and efficiently. For our methods, we first introduce the concept of self-consistency trees, which model multi-step transformations as a tree structure. Based on this, we define a consistency score using a self-consistency forest to enable robust evaluation across multiple tasks. Experimental results show that our consistency metric reliably estimates model performance for both translation and programming tasks, and closely matches WMT benchmark metrics such as AutoRank and CometKiwi for machine translation.

\section*{Limitations}
While this study offers valuable insights into LLM consistency via the ConsistencyChecker framework, it has several limitations. The evaluation is limited to machine translation and code generation, excluding tasks like arithmetic. It primarily covers only 8 models in mainstream model families. Additionally, the study relies solely on automated metrics such as embedding similarity and BLEU, which may not fully reflect human judgment, especially for tasks requiring creativity or nuanced interpretation. Expanding task coverage, model diversity, and incorporating human evaluation in future work would strengthen the generalizability and robustness of the framework.

\bibliography{custom}

\clearpage
\appendix
\label{sec:appendix}

\section{Potential Risks}
The potential risks associated with this work include the misuse of the generated artifacts outside of research contexts. To mitigate this, we have included a license in the appendix that specifies the intended use of the artifacts and restricts their use to research purposes only. We release them under the MIT license: \url{https://github.com/ulab-uiuc/consistencychecker}.

\section{Artifact Use}
\paragraph{License compliance}
We confirm that our use of existing artifacts complies with their respective licenses. For artifacts created in this work, we explicitly specify intended use in the documentation, ensuring compatibility with original access conditions. Derivatives of data accessed for research purposes are restricted to research contexts only.

\paragraph{Model licenses}
Claude-3.5-Sonnet, GPT-4, GPT-4o-mini, and Gemini-1.5-Pro are all proprietary LLMs.
In the Qwen family, Qwen-2.5-1.5B, Qwen-2.5-7B, Qwen-2.5-14B, and Qwen-2.5-32B are licensed under Apache-2.0, while Qwen-2.5-72B uses the Qwen license.
Phi-3-Medium is under the MIT license.
Mistral-Large is under the Mistral Research License.
LLaMA-3-70B is licensed under the llama3 license, while LLaMA-3.1-8B and LLaMA-3.1-79B are under the llama3.1 license.
nvidia/NV-Embed-v2, the embedding model we used, is under the license CC-BY-NC 4.0.

\paragraph{API compliance} Open-source LLMs greater than 8B are accessed through Fireworks AI API. The Claude-3.5-Sonnet is accessed via official Anthropic endpoints. Gemini-1.5-Pro is visited from Gemini official API endpoints, and so is Mistral-Large, which is from Mistral official API. Phi-3-Medium is accessed through the NVIDIA Nim API. We accept and are compliant to each of their terms, respectively.

\paragraph{Benchmark use}
We do not use any existing benchmarks directly; we only reference their published results, and also the language pairs used in WMT 2024 automatic benchmarks.

\section{Documentation of Artifacts}
We provide comprehensive documentation for the artifacts used and created in this work. This includes details on the domains, languages, and linguistic phenomena covered, as well as the demographic groups represented. This study involves the generation of benchmarks and evaluation log files involving English, German, French, Spanish, Czech, Ukrainian, Japanese, and Chinese. 

\section{Statistics for Data}
We report relevant statistics for the data used and created in this work. The four evaluator models involved in this study each generated 2 benchmark files, with one for the translation task and one for the AI-assisted programming task. Each of these benchmark files contains 10 self-consistent root nodes. For each LLM evaluated on each of these two tasks, it generated a self-consistency tree for each root node, and the degree and height of which are described in the paper.

\section{Model Size and Budget}
\paragraph{Model size and configuration}
We report the number of parameters for each model used in this study:

Claude-3.5-Sonnet: unknown size;

GPT-4: unknown size;

GPT-4o-mini: unknown size;

Gemini-1.5-Pro: unknown size;

Mistral-Large: 128 billion parameters;

LLaMA-3-70B: 70 billion parameters;

Phi-3-Medium: 14 billion parameters;

Qwen-2.5-1.5B: 1.5 billion parameters;

Qwen-2.5-7B: 7 billion parameters;

Qwen-2.5-14B: 14 billion parameters;

Qwen-2.5-32B: 32.5 billion parameters;

Qwen-2.5-72B: 72 billion parameters;

LLaMA-3.1-8B: 8 billion parameters;

LLaMA-3.1-70B: 70 billion parameters.

\paragraph{Computational resources}
This project was conducted on 2 NVIDIA RTX A6000 GPUs. All LLMs with fewer than 8 billion parameters were hosted locally using vLLM on these GPUs, with random seed set to 42. Larger models and proprietary APIs (such as GPT-4o-mini, Qwen-2.5-14B, Qwen-2.5-70B, and LLaMA-3.1-72B) were accessed through external APIs.

\paragraph{Model variants}
All LLMs in the LLaMA and Qwen families used in this work are instruction-tuned versions. No additional quantization was applied to any model.

\section{Experimental Settings}
We discuss the experimental setup, including hyperparameter search and the best-found hyperparameter values. The temperature for text generation was set to 0.6. The tree structure used for evaluation, on the two tasks, aside from the experiment for comparison with WMT 2024,  had built 10 self-consistency tree with an out-degree of 3 (operation pairs) and a height of 3. These parameters were chosen based on preliminary experiments to balance diversity and consistency in the generated outputs. In the experiments to compare our framework with WMT 2024, for language pairs with English as one of the languages, the out-degree is 1 and the height of the trees is 12. For all those pairs withoiut English as one of the two languages, the out-degree is 3 and the height is 3.

\section{Parameters for Packages}
We used the following packages and reported their implementation, model, and parameter settings: \texttt{nltk.translate.bleu\_score}: For calculating BLEU scores, we used the \texttt{sentence\_bleu} function with \texttt{SmoothingFunction().method1}.

\section{Information about Use of AI Assistants}
We acknowledge the use of AI assistants in the preparation of this work. The AI assistants were used for generating and refining text, as well as for providing suggestions on experimental design and hyperparameter tuning. All outputs generated by AI assistants were reviewed and validated by the authors to ensure accuracy and relevance.

\begin{table*}[t!]
\begin{center}
\begin{minipage}{0.48\textwidth}
\centering
\begin{small}
\caption{\textbf{Machine translation consistency evaluation with different path lengths.} Scores(\%) evaluate LLMs' ability to attain semantics after translating across different languages.}
\scalebox{0.95}{
\begin{tabular}{lccc}
\toprule
\textbf{Evaluatee} & \textbf{$C_1(\mathcal{F})$} & \textbf{$C_2(\mathcal{F})$} & \textbf{$C_3(\mathcal{F})$} \\
\midrule
GPT-4o-mini & \textbf{99.2}$_{\pm0.0}$ & \textbf{98.7}$_{\pm0.0}$ & \textbf{98.0}$_{\pm0.0}$ \\
\midrule
Qwen-2.5-1.5B & 91.9$_{\pm0.1}$ & 86.5$_{\pm0.1}$ & 80.3$_{\pm0.5}$ \\
Qwen-2.5-7B & 96.4$_{\pm0.1}$ & 93.5$_{\pm0.3}$ & 90.0$_{\pm0.8}$ \\
Qwen-2.5-14B & 97.4$_{\pm0.0}$ & 96.1$_{\pm0.1}$ & 94.7$_{\pm0.1}$ \\
Qwen-2.5-32B & 97.6$_{\pm0.1}$ & 97.4$_{\pm0.0}$ & 96.4$_{\pm0.0}$ \\
Qwen-2.5-72B & 99.1$_{\pm0.0}$ & 98.3$_{\pm0.0}$ & 97.2$_{\pm0.0}$ \\
\midrule
LLaMA-3.1-8B & 89.5$_{\pm0.1}$ & 75.7$_{\pm1.6}$ & 67.5$_{\pm3.0}$ \\
LLaMA-3.1-70B & 87.4$_{\pm0.7}$ & 80.7$_{\pm1.1}$ & 71.9$_{\pm3.2}$ \\
\bottomrule
\end{tabular}
}
\label{tab:translation}
\end{small}
\end{minipage}
\hfill
\begin{minipage}{0.48\textwidth}
\centering
\begin{small}
\caption{\textbf{AI-assisted programming consistency evaluation with different path lengths.} Scores(\%) evaluate LLMs' ability to preserve functionality after multi-step algorithm transformation.}
\scalebox{0.95}{
\begin{tabular}{lccc}

\toprule
\textbf{Evaluatee} & \textbf{$C_1(\mathcal{F})$} & \textbf{$C_2(\mathcal{F})$} & \textbf{$C_3(\mathcal{F})$} \\
\midrule
GPT-4o-mini & 90.6$_{\pm0.2}$ & 84.7$_{\pm0.6}$ & 76.5$_{\pm2.7}$ \\
\midrule
Qwen-2.5-1.5B & 90.2$_{\pm0.2}$ & 80.0$_{\pm0.6}$ & 63.4$_{\pm1.4}$ \\
Qwen-2.5-7B & 85.4$_{\pm0.2}$ & 80.2$_{\pm0.2}$ & 71.7$_{\pm0.4}$ \\
Qwen-2.5-14B & 90.3$_{\pm0.2}$ & 83.1$_{\pm0.6}$ & 79.9$_{\pm1.0}$ \\
Qwen-2.5-32B & \textbf{91.1}$_{\pm0.4}$ & \textbf{88.3}$_{\pm0.7}$ & \textbf{85.1}$_{\pm1.1}$ \\
Qwen-2.5-72B & 90.8$_{\pm0.2}$ & 85.3$_{\pm0.5}$ & 77.0$_{\pm1.9}$ \\
\midrule
LLaMA-3.1-8B & 88.1$_{\pm0.2}$ & 76.0$_{\pm0.5}$ & 60.4$_{\pm1.0}$ \\
LLaMA-3.1-70B & 87.9$_{\pm0.4}$ & 82.7$_{\pm0.6}$ & 83.5$_{\pm1.0}$ \\
\bottomrule
\end{tabular}
}
\label{tab:code}
\end{small}
\end{minipage}
\end{center}
\vspace{-2mm}
\end{table*}

\section{Additional Experimental Results}

\paragraph{Additional results on path length} We also release our results on these 2 defined tasks in these tables: Table~\ref{tab:translation}, \ref{tab:code}, and~\ref{combined-table-translation-and-code}, where we report the numbers for $C_1(\mathcal{F})$, $C_2(\mathcal{F})$, and $C_3(\mathcal{F})$. They correspond to and provide more information than the tables Table~\ref{combined-table-translation-and-code-l3} and~\ref{sample-table-merged-bleu-evaluation}. They also manifest the trend that under the same circumstances $C_2(\mathcal{F})$ is lower than $C_1(\mathcal{F})$, and $C_3(\mathcal{F})$ is lower than $C_2(\mathcal{F})$.

\label{appendix:correlation-study}

\begin{figure*}[ht!]
    \centering
    \begin{minipage}[t]{0.48\linewidth}
        \includegraphics[width=\linewidth]{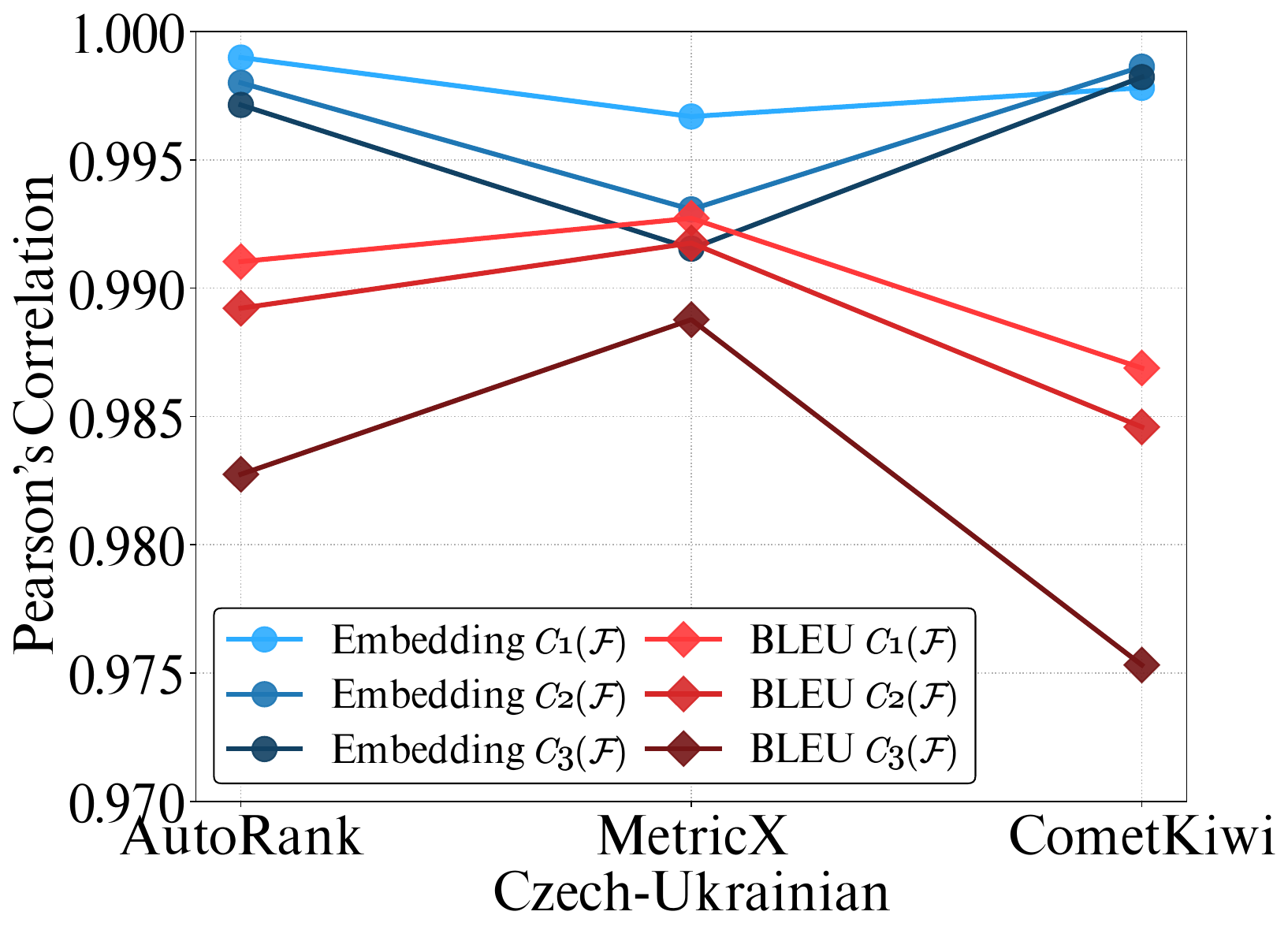}
        \includegraphics[width=\linewidth]{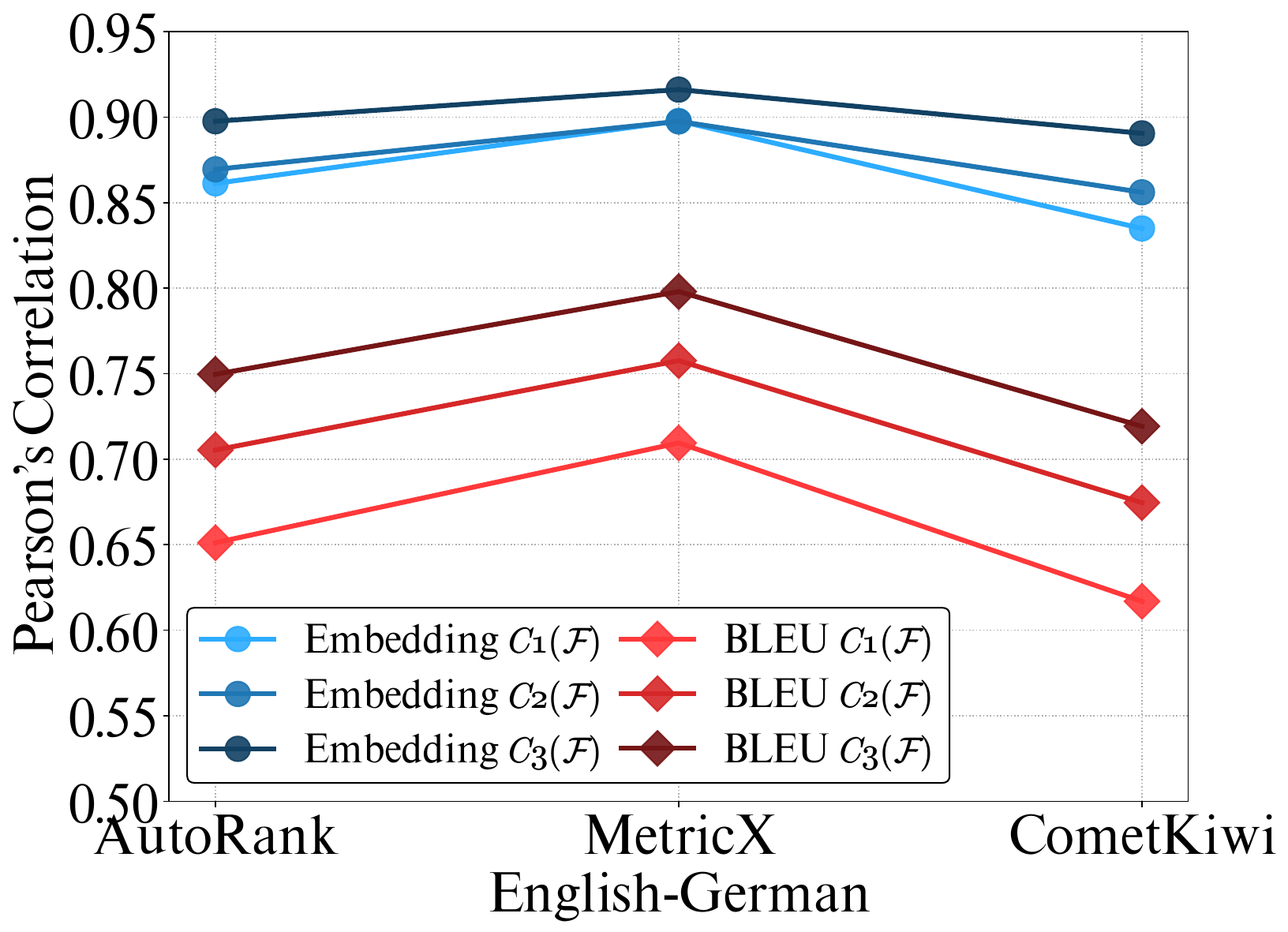}
        \includegraphics[width=\linewidth]{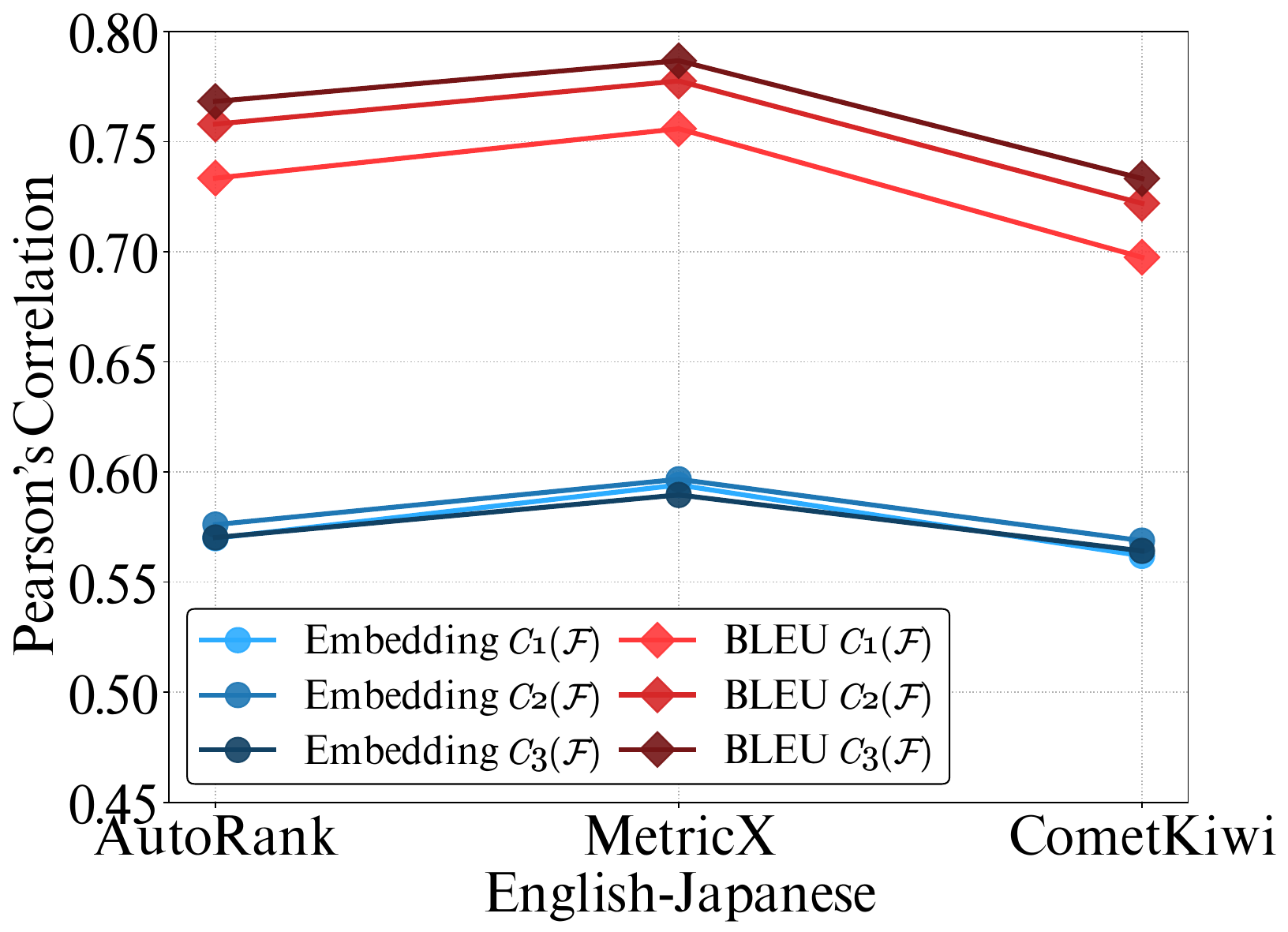}
    \end{minipage}
    \hfill
    \begin{minipage}[t]{0.48\linewidth}
        \includegraphics[width=\linewidth]{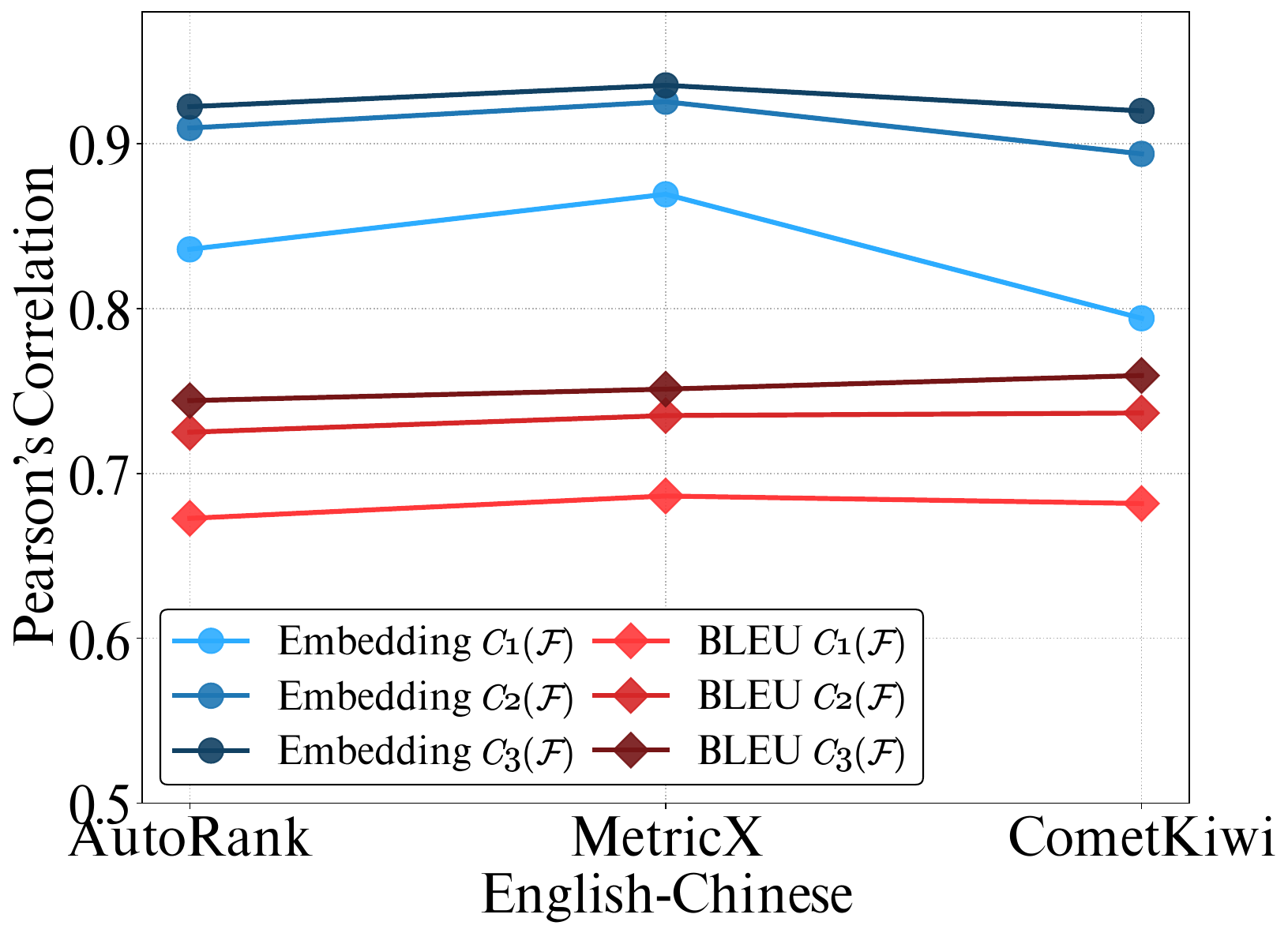}
        \includegraphics[width=\linewidth]{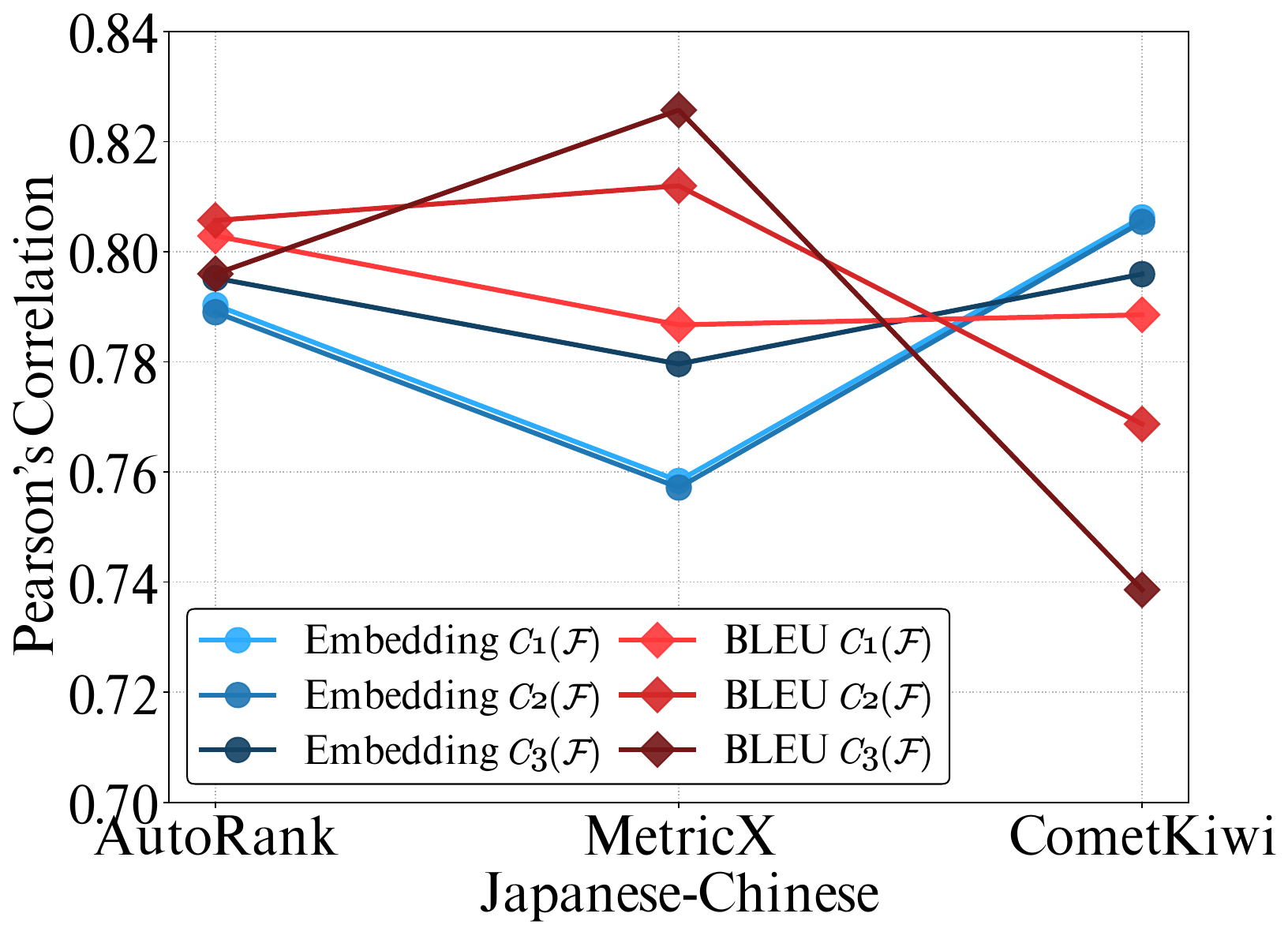}
    \end{minipage}
    
    \caption{\textbf{Correlation between ConsistencyChecker evaluation and metrics in WMT 2024 benchmarks}. The three figures on the left column from top to bottom correspond to the Czech-Ukrainian, English-German, and English-Japanese language pairs. The two figures on the right column from top to bottom correspond to the English-Chinese and Japanese-Chinese language pairs. Each of these figures shows the correlation between the ConsistencyChecker scores by embedding and by BLEU at $C_1(\mathcal{F})$, $C_2(\mathcal{F})$, and $C_3(\mathcal{F})$, with the three metrics used in WMT 2024, AutoRank, MetricX, and CometKiwi. The results show that both embedding-based and BLEU-based ConsistencyChecker scores share high correlation with the three metrics in WMT 2024.}
    \label{fig:correlation}
\end{figure*}

\paragraph{Additional results on correlation scores in WMT 2024}
Table~\ref{consistency-vs-wmt-translation-2} shows the comparison between our ConsistencyChecker rankings and WMT 2024 rankings. In most cases, at least 4 out of 6 LLMs share the same order across both evaluations. To further quantify the relationship, we calculate Pearson’s correlation between our consistency scores and WMT’s three automated metrics. Since AutoRank and MetricX are inversely related to performance, we use their absolute values for the correlation analysis. As illustrated in Figure~\ref{fig:correlation}, there is a strong correlation between ConsistencyChecker scores and WMT 2024 evaluations. This relationship is evident using both NV-Embed-v2-based cosine embedding similarity and BLEU as the similarity metric.

\begin{table*}[t]
  \caption{\textbf{Consistency results under different evaluator models and different path lengths.} (Top) Machine translation consistency scores (\%) across model scales (7B–72B) and transform complexity ($C_1(\mathcal{F})$–$C_3(\mathcal{F})$). (Bottom) Code equivalence preservation scores (\%) for algorithm transformations at varying abstraction levels ($C_1(\mathcal{F})$–$C_3(\mathcal{F})$).}
  \vspace{-2mm}
  \label{combined-table-translation-and-code}
  \begin{center}
    \begin{small}
      \setlength{\tabcolsep}{4pt}
      \begin{tabular}{l|ccc|ccc}
        \toprule
        \multirow{2}{*}{\diagbox{Evaluatee}{Evaluator}}
          & \multicolumn{3}{c|}{\textbf{Qwen-2.5-7B}}
          & \multicolumn{3}{c}{\textbf{Qwen-2.5-72B}} \\
          & $C_1(\mathcal{F})$ & $C_2(\mathcal{F})$ & $C_3(\mathcal{F})$ & $C_1(\mathcal{F})$ & $C_2(\mathcal{F})$ & $C_3(\mathcal{F})$ \\
        \midrule
        \multicolumn{7}{c}{\textbf{Machine translation}} \\
        \midrule
        Qwen-2.5-7B    & \textbf{90.1}$_{\pm0.9}$ & 82.8$_{\pm3.0}$ & 76.8$_{\pm5.1}$
                       & 96.4$_{\pm0.1}$ & 93.5$_{\pm0.3}$ & 90.0$_{\pm0.8}$ \\
        Qwen-2.5-72B   & 90.0$_{\pm1.8}$         & \textbf{84.4}$_{\pm4.2}$ & \textbf{78.2}$_{\pm7.3}$
                       & \textbf{99.1}$_{\pm0.0}$ & \textbf{98.3}$_{\pm0.0}$ & \textbf{97.2}$_{\pm0.0}$ \\
        \midrule
        LLaMA-3.1-8B   & 81.6$_{\pm0.5}$         & 73.0$_{\pm1.3}$         & 61.7$_{\pm4.8}$
                       & 89.5$_{\pm0.1}$         & 75.7$_{\pm1.6}$         & 67.5$_{\pm3.0}$ \\
        LLaMA-3.1-70B  & 82.3$_{\pm1.8}$         & 68.8$_{\pm2.3}$         & 30.8$_{\pm9.2}$
                       & 87.4$_{\pm0.7}$         & 80.7$_{\pm1.1}$         & 71.9$_{\pm3.2}$ \\
        \midrule
        \multicolumn{7}{c}{\textbf{AI-assisted programming}} \\
        \midrule
        Qwen-2.5-7B    & 86.2$_{\pm0.4}$         & 78.2$_{\pm0.8}$         & 72.2$_{\pm1.9}$
                       & 85.4$_{\pm0.2}$         & 80.2$_{\pm0.2}$         & 71.7$_{\pm0.4}$ \\
        Qwen-2.5-72B   & 87.2$_{\pm0.2}$         & 80.1$_{\pm0.6}$         & 79.4$_{\pm2.3}$
                       & \textbf{90.8}$_{\pm0.2}$ & \textbf{85.3}$_{\pm0.5}$ & 77.0$_{\pm1.9}$ \\
        \midrule
        LLaMA-3.1-8B   & 84.4$_{\pm0.2}$         & 72.1$_{\pm0.4}$         & 61.0$_{\pm1.1}$
                       & 88.1$_{\pm0.2}$         & 76.0$_{\pm0.5}$         & 60.4$_{\pm1.0}$ \\
        LLaMA-3.1-70B  & \textbf{91.0}$_{\pm0.5}$ & \textbf{85.6}$_{\pm1.0}$ & \textbf{81.5}$_{\pm2.1}$
                       & 87.9$_{\pm0.4}$         & 82.7$_{\pm0.6}$         & \textbf{83.5}$_{\pm1.0}$ \\
        \bottomrule
      \end{tabular}
    \end{small}
  \end{center}
  \vskip -0.1in
\end{table*}

\begin{table*}[t]
    \caption{\textbf{Comprehensive results on machine translation tasks.} Consistency scores (\%) across model and transform complexity ($C_1(\mathcal{F})$-$C_3(\mathcal{F})$) using embedding similarity and BLEU scores, versus metrics in WMT 2024 benchmarks. The rows are arranged using WMT rankings.}
    \vspace{-2mm}
    \label{consistency-vs-wmt-translation-2}
    \begin{center}
    \begin{small}
    \scalebox{0.9}{
    \begin{tabular}{lccccccccc}
    \toprule
        \multirow{2}{*}{\textbf{Czech-Ukrainian}} & \multicolumn{3}{c}{Embedding} & \multicolumn{3}{c}{BLEU} & \multirow{2}{*}{AutoRank$\downarrow$} & \multirow{2}{*}{MetricX$\downarrow$} & \multirow{2}{*}{CometKiwi$\uparrow$}\\
        & $C_1(\mathcal{F})$ & $C_2(\mathcal{F})$ & $C_3(\mathcal{F})$ & $C_1(\mathcal{F})$ & $C_2(\mathcal{F})$ & $C_3(\mathcal{F})$ \\
    \midrule
    Claude-3.5-Sonnet & 99.4 & 98.9 & 98.1 & 87.1 & 80.4 & 67.3 & 1.7 & 1.0 & 0.683\\
    GPT-4 & 98.7 & 97.8 & 96.4 & 78.1 & 69.2 & 55.5 & 2.0 & 1.4 & 0.677\\
    Gemini-1.5-Pro & 99.2 & 98.5 & 97.5 & 86.0 & 78.3 & 67.6 & 2.0 & 1.2 & 0.668\\
    Mistral-Large & 98.6 & 97.6 & 96.5 & 78.3 & 68.8 & 56.7 & 2.3 & 1.6 & 0.666\\
    LLaMA-3-70B & 96.4 & 96.7 & 95.9 & 79.2 & 70.4 & 55.8 & 2.6 & 2.0 & 0.661\\
    Phi-3-Medium & 71.1 & 58.3 & 44.9 & 28.9 & 14.6 & 5.6 & 9.1 & 6.5 & 0.425\\
    \midrule
        \multirow{2}{*}{\textbf{English-German}} & \multicolumn{3}{c}{Embedding} & \multicolumn{3}{c}{BLEU} & \multirow{2}{*}{AutoRank$\downarrow$} & \multirow{2}{*}{MetricX$\downarrow$} & \multirow{2}{*}{CometKiwi$\uparrow$}\\
        & $C_1(\mathcal{F})$ & $C_2(\mathcal{F})$ & $C_3(\mathcal{F})$ & $C_1(\mathcal{F})$ & $C_2(\mathcal{F})$ & $C_3(\mathcal{F})$ \\
    \midrule
    GPT-4 & 99.4 & 99.2 & 99.1 & 85.8 & 83.5 & 82.4 & 1.8 & 1.4 & 0.700\\
    Claude-3.5-Sonnet & 99.8 & 99.8 & 99.8 & 95.6 & 94.8 & 94.4 & 1.9 & 1.4 & 0.695\\
    Mistral-Large & 99.4 & 99.3 & 99.2 & 88.5 & 86.9 & 85.5 & 2.0 & 1.5 & 0.694\\
    Gemini-1.5-Pro & 99.8 & 99.7 & 99.6 & 93.3 & 92.8 & 91.4 & 2.2 & 1.5 & 0.688\\
    LLaMA-3-70B & 99.3 & 99.1 & 98.9 & 88.4 & 86.2 & 84.3 & 2.5 & 1.7 & 0.686\\
    Phi-3-Medium & 98.5 & 97.9 & 97.2 & 81.2 & 76.2 & 72.7 & 3.4 & 2.0 & 0.657\\
    \midrule
        \multirow{2}{*}{\textbf{English-Japanese}} & \multicolumn{3}{c}{Embedding} & \multicolumn{3}{c}{BLEU} & \multirow{2}{*}{AutoRank$\downarrow$} & \multirow{2}{*}{MetricX$\downarrow$} & \multirow{2}{*}{CometKiwi$\uparrow$}\\
        & $C_1(\mathcal{F})$ & $C_2(\mathcal{F})$ & $C_3(\mathcal{F})$ & $C_1(\mathcal{F})$ & $C_2(\mathcal{F})$ & $C_3(\mathcal{F})$ \\
    \midrule
    Claude-3.5-Sonnet & 99.4 & 99.2 & 99.0 & 89.2 & 87.3 & 85.8 & 1.5 & 2.3 & 0.744\\
    Gemini-1.5-Pro & 99.3 & 99.2 & 99.1 & 88.1 & 86.3 & 84.9 & 1.7 & 2.5 & 0.734\\
    GPT-4 & 98.4 & 98.1 & 97.8 & 76.6 & 72.2 & 69.4 & 1.7 & 2.7 & 0.740\\
    LLaMA-3-70B & 93.5 & 90.5 & 87.9 & 71.4 & 65.4 & 60.0 & 2.6 & 3.5 & 0.714\\
    Phi-3-Medium & 97.4 & 96.1 & 95.3 & 65.5 & 57.1 & 51.8 & 2.8 & 3.6 & 0.709\\
    Mistral-Large & 98.2 & 97.4 & 96.9 & 79.2 & 73.9 & 70.7 & 2.9 & 3.8 & 0.707\\
    \midrule
        \multirow{2}{*}{\textbf{English-Chinese}} & \multicolumn{3}{c}{Embedding} & \multicolumn{3}{c}{BLEU} & \multirow{2}{*}{AutoRank$\downarrow$} & \multirow{2}{*}{MetricX$\downarrow$} & \multirow{2}{*}{CometKiwi$\uparrow$}\\
        & $C_1(\mathcal{F})$ & $C_2(\mathcal{F})$ & $C_3(\mathcal{F})$ & $C_1(\mathcal{F})$ & $C_2(\mathcal{F})$ & $C_3(\mathcal{F})$ \\
    \midrule
    Claude-3.5-Sonnet & 99.5 & 99.4 & 99.3 & 89.8 & 88.7 & 86.9 & 1.7 & 3.0 & 0.703\\
    Gemini-1.5-Pro & 99.4 & 99.3 & 99.3 & 89.4 & 88.5 & 86.9 & 1.8 & 3.1 & 0.698\\
    GPT-4 & 98.6 & 98.3 & 98.1 & 76.4 & 73.3 & 71.4 & 2.0 & 3.3 & 0.693\\
    Mistral-Large & 97.5 & 97.7 & 97.5 & 82.5 & 79.5 & 77.6 & 2.8 & 4.0 & 0.665\\
    LLaMA-3-70B & 98.6 & 98.3 & 97.9 & 82.4 & 78.8 & 75.9 & 2.8 & 3.9 & 0.662\\
    Phi-3-Medium & 98.1 & 97.3 & 96.6 & 72.2 & 65.9 & 61.8 & 3.1 & 4.2 & 0.648\\
    \midrule
    \multirow{2}{*}{\textbf{Japanese-Chinese}} & \multicolumn{3}{c}{Embedding} & \multicolumn{3}{c}{BLEU} & \multirow{2}{*}{AutoRank$\downarrow$} & \multirow{2}{*}{MetricX$\downarrow$} & \multirow{2}{*}{CometKiwi$\uparrow$}\\
    & $C_1(\mathcal{F})$ & $C_2(\mathcal{F})$ & $C_3(\mathcal{F})$ & $C_1(\mathcal{F})$ & $C_2(\mathcal{F})$ & $C_3(\mathcal{F})$ \\
    \midrule
    Claude-3.5-Sonnet & 98.7 & 97.7 & 96.0 & 77.4 & 65.4 & 49.6 & 1.7 & 3.5 & 0.603 \\
    Gemini-1.5-Pro & 98.8 & 98.1 & 96.8 & 80.1 & 70.0 & 54.0 & 1.9 & 3.5 & 0.595 \\
    GPT-4 & 97.6 & 96.0 & 93.7 & 65.3 & 51.6 & 37.3 & 2.1 & 3.8 & 0.597 \\
    LLaMA-3-70B & 96.2 & 93.4 & 88.9 & 64.6 & 48.0 & 28.5 & 3.1 & 4.7 & 0.578 \\
    Mistral-Large & 98.1 & 96.7 & 94.4 & 69.4 & 55.6 & 39.9 & 3.5 & 4.9 & 0.568 \\
    Phi-3-Medium & 93.9 & 89.3 & 84.9 & 52.1 & 38.6 & 26.0 & 4.0 & 5.1 & 0.552\\
    \bottomrule
    \end{tabular}
    }
    \end{small}
    \end{center}
    \vskip -0.1in
\end{table*}

\begin{figure*}[t!]
\centering
\begin{tcolorbox}[colback=blue!2!white, colframe=gray!50!blue]
\begin{minipage}{\textwidth}
\textbf{Root Node:}
def main():
    return "In the rapidly evolving landscape of artificial intelligence (AI), the integration of these advanced technologies into healthcare systems represents a significant frontier with profound implications for medical diagnostics and patient care. Recent advancements have demonstrated the potential of AI to enhance diagnostic accuracy, streamline clinical workflows, and personalize treatment plans, thereby improving patient outcomes and reducing healthcare costs. However, the deployment of AI in this sensitive domain is not without its challenges. Ethical considerations, such as data privacy, algorithmic bias, and the potential for job displacement, are paramount. Moreover, the regulatory landscape must adapt to ensure that AI applications meet stringent safety and efficacy standards. Despite these hurdles, the collaboration between tech developers, healthcare providers, and regulatory bodies is fostering a cautious yet optimistic approach to AI integration. The future of healthcare may well be defined by the seamless fusion of human expertise and machine intelligence, creating a hybrid model that leverages the strengths of both to deliver more effective and efficient care."
\end{minipage}
\end{tcolorbox}

\begin{tcolorbox}[colback=red!2!white, colframe=gray!50!red]
\begin{minipage}{\textwidth}
\textbf{LLaMA-3.1-8B Leaf Node:}
def main():
    return "In the constantly evolving scenario of artificial intelligence (AI), the integration of these advanced technologies in medical systems represents a significant frontier with important implications for medical diagnosis and patient care. Recent advances have demonstrated the potential of AI to improve diagnostic accuracy, simplify clinical workflows, and personalize treatment plans, leading to better patient outcomes and a reduction in medical costs. However, implementing AI in this sensitive domain does not come without its challenges. Ethical considerations, such as data privacy, algorithmic bias, and job displacement, are fundamental. Furthermore, the regulatory landscape must adapt to ensure that AI applications meet strict standards of safety and efficacy. Despite these difficulties, collaboration between technology developers, healthcare providers, and regulatory bodies is promoting a cautious yet optimistic approach to the integration of AI. The future of healthcare may be defined by the seamless fusion of human experience and artificial intelligence, creating a hybrid model that leverages the strengths of both to offer more effective and efficient care."
\end{minipage}
\end{tcolorbox}

\begin{tcolorbox}[colback=green!2!white, colframe=gray!50!green]
\begin{minipage}{\textwidth}
\textbf{LLaMA-3.1-70B Leaf Node:}
def main():
    return "In the rapidly evolving landscape of artificial intelligence (AI), the integration of these advanced technologies into healthcare systems represents a significant frontier with profound implications for medical diagnosis and patient care. Recent advances have demonstrated AI's potential to improve diagnostic accuracy, streamline clinical workflows, and personalize treatment plans, thereby enhancing patient outcomes and reducing healthcare costs. However, the deployment of AI in this sensitive domain is not without challenges. Ethical considerations, such as data privacy, algorithmic bias, and potential job displacement, are paramount. Moreover, the regulatory landscape must adapt to ensure that AI applications meet stringent safety and efficacy standards. Despite these obstacles, collaboration between technology developers, healthcare providers, and regulatory agencies is fostering a cautious yet optimistic approach to AI integration. The future of healthcare may be defined by the seamless fusion of human expertise and machine intelligence, creating a hybrid model that leverages the strengths of both to deliver more effective and efficient care."
\end{minipage}
\end{tcolorbox}
\caption{\textbf{Example of root nodes and leaf nodes in a self-consistency tree}.}
\label{trio}
\end{figure*}

\section{Additional Case Study}
\label{example-case-study}

\subsection{Full Examples of Leaf Nodes}
We include the full example of one root node and two leaf nodes generated by LLaMA-3.1-8B and LLaMA-3.1-70B for consistency evaluation in Figure~\ref{trio}.

\subsection{Full Examples of Self-consistency Tree}
In this section, we present two complete self-consistency trees with a depth of 3 and a branching factor of 3 for the machine translation task. These trees are generated by two different models: Qwen-2.5-7B and Qwen-2.5-72B.

This is the full content of the self-consistency tree generated by Qwen-2.5-7B on machine translation tasks.

\vspace{2mm}
{\ttfamily\small
\xhdr{Root (Level 0, NV-Embed-v2 similarity to root 1.0000)} In the ever-evolving landscape of global economic governance, the recent policy
adjustments by the European Central Bank (ECB) stand out as a significant marker

of the shifting tides in international finance. The ECB's decision to
recalibrate its quantitative easing (QE) program, alongside a series of nuanced
interest rate adjustments, has sent shockwaves through the financial markets,
particularly affecting emerging economies that are intricately linked to the
\begin{lstlisting}
Eurozone's economic health. These changes, while designed to stabilize the Euro
\end{lstlisting}

and stimulate economic growth within the European Union, have had far-reaching
consequences, including capital flight, currency volatility, and heightened risk
aversion among investors. The ECB's forward guidance, which aims to provide
clarity and predictability, has been met with mixed reactions, as some market
analysts argue that it lacks the necessary flexibility to address the diverse
and dynamic economic conditions across different regions. In response, smaller
financial institutions and central banks in emerging markets have been compelled
to adopt a range of adaptive measures, from tightening monetary policies to
diversifying their investment portfolios, in an effort to insulate their
economies from external shocks and capitalize on the emerging opportunities. The
interplay between these global and local economic strategies underscores the
complex and interconnected nature of the modern financial system, where the
decisions of one major player can have profound implications for the rest of the
world.

\xhdr{Node 0 (Level 1, NV-Embed-v2 similarity to root 0.3336)} The communication and forward strategies between global and local economic
strategies highlight the complexity and interconnectivity of modern financial
systems, where one player's decisions can have far-reaching impacts on the rest
of the world.

\xhdr{Node 0-0 (Level 2, NV-Embed-v2 similarity to root 0.2802)} The communication and promotion strategy between global and local economic
strategies highlights the complexity and interconnection of modern financial
systems, where the decisions of a single actor can have reverberating impacts on
the rest of the world.

\xhdr{Node 0-0-0 (Level 3, NV-Embed-v2 similarity to root 0.2820)} The communication and promotion strategies between global and local economic
strategies highlight the complexity and interconnectedness of modern financial
systems, where the decisions of a single actor can have repercussions on the
rest of the world.

\xhdr{Node 0-0-1 (Level 3, NV-Embed-v2 similarity to root 0.2400)} The communications and propaganda strategy between global and local economic
strategies illustrates the complexity and the internal connection of modern
financial systems, where the decisions of a single actor can affect the rest of
the world.

\xhdr{Node 0-0-2 (Level 3, NV-Embed-v2 similarity to root 0.2693)} The communication and promotion strategy between global and local economic
strategies highlights the complexity and interconnection of modern financial
systems, where the decisions of a single actor can have resonant effects on the
rest of the world.

\xhdr{Node 0-1 (Level 2, NV-Embed-v2 similarity to root 0.2774)} The communication and foresight strategies between global and local business
strategies illustrate the complexity and interdependence of modern financial
systems, where the decisions of one player can have far-reaching effects on the
rest of the world.

\xhdr{Node 0-1-0 (Level 3, NV-Embed-v2 similarity to root 0.2717)} The communication and foresight strategy between global and local business
strategies illustrates the complexity and interdependence of modern financial
systems, where the decisions of one player can have repercussions on the rest of
the world.

\xhdr{Node 0-1-1 (Level 3, NV-Embed-v2 similarity to root 0.2615)} The communication and prediction strategies between global and regional business
concepts illustrate the complexity and interconnectedness of modern financial
systems, where the decisions of one player can have far-reaching effects on the
rest of the world.

\xhdr{Node 0-1-2 (Level 3, NV-Embed-v2 similarity to root 0.2721)} The communication and forecast strategy between global and local business
strategies illustrate the complexity and interdependence of modern financial
systems, where the decisions of one player can have long-lasting effects on the
rest of the world.

\xhdr{Node 0-2 (Level 2, NV-Embed-v2 similarity to root 0.2783)} The strategy of communication and advancement between global and local
strategies highlights the complexity and interconnection of modern financial
systems, where the decisions of a single player can have long-term repercussions
for the rest of the world.

\xhdr{Node 0-2-0 (Level 3, NV-Embed-v2 similarity to root 0.2595)} The communication and advancement strategy between global and local strategies
highlights the complexity and interconnection of modern financial systems, where
the decisions of a single player can have long-term repercussions for the rest
of the world.

\xhdr{Node 0-2-1 (Level 3, NV-Embed-v2 similarity to root 0.2852)} The strategy of communication and progress balancing between global and local
strategies illustrates the complexity and connection of modern financial
systems, where the decisions of a single player can have long-term effects on
the rest of the world.

\xhdr{Node 0-2-2 (Level 3, NV-Embed-v2 similarity to root 0.2687)} The strategy of communication and advancement between global and local
strategies highlights the complexity and interconnection of modern financial
systems, in which the decisions of a single player can have long-term
repercussions for the rest of the world.

\xhdr{Node 1 (Level 1, NV-Embed-v2 similarity to root 0.9537)} In the constantly changing landscape of global economic policy, the recent
measures of the European Central Bank (ECB) stand out as a significant sign of
evolving directions in the international financial world. The ECB's decision to
adjust its Quantitative Easing (QE) programs and a series of nuanced interest
rate adjustments have sparked shockwaves in financial markets, particularly in
emerging economies closely linked to the economic health of the eurozone. These
changes, aimed at stabilizing the euro and fostering growth in the European
Union, have far-reaching consequences, including capital flight, currency
volatility, and heightened risk aversion among investors. The forward guidance
of the ECB, which seeks to provide clarity and predictability, is met with mixed
reactions, as some market analysts argue that it lacks the necessary flexibility
to accommodate the diverse and dynamic economic conditions in different regions.
In response, smaller financial institutions and central banks in emerging
economies are forced to take a range of adaptive measures, from tightening their
interest conditions to diversifying their investment portfolios, to shield their
economies from external shocks and capitalize on emerging opportunities. The
interaction between these global and local economic strategies underscores the
complex and interconnected nature of modern financial systems, where the
decisions of a major player can have profound impacts on the rest of the world.

\xhdr{Node 1-0 (Level 2, NV-Embed-v2 similarity to root 0.9234)} In the constantly evolving global economic landscape, the recent measures of the
European Central Bank (ECB) stand out as a significant sign of emerging
directions in the international financial world. The ECB's decision to modulate
its Quantitative Easing (QE) programs and a series of subtle adjustments to
interest rates have had repercussions in financial markets, particularly in
emerging economies closely linked to the economic health of the eurozone. These
changes, aimed at stabilizing the euro and promoting growth in the European
Union, have consequences far beyond, including capital flight, currency
volatility, and an increased risk aversion among investors. The ECB's forward
guidance, which aims to provide clarity and predictability, is met with mixed
reactions, as some market analysts suggest that it lacks the flexibility needed
to adapt to diverse and dynamic economic conditions in different regions. In
response, smaller financial institutions and central banks of emerging economies
are forced to adopt a range of adaptive measures, from reducing interest rates
to diversifying investment portfolios, in order to protect their economies from
external shocks and seize emerging opportunities. The interaction between these
global and local economic strategies highlights the complex and interconnected
nature of modern financial systems, where the decisions of a major player can
have profound impacts on the rest of the world.

\xhdr{Node 1-0-0 (Level 3, NV-Embed-v2 similarity to root 0.0520)} None

\xhdr{Node 1-0-1 (Level 3, NV-Embed-v2 similarity to root 0.8577)} In the constantly changing global economic landscape, the recent measures of the
European Central Bank (ECB) represent an important signal for the emerging
directions on the international financial market. The decision of the ECB to
modulate its Quantitative Liquidity Provision (QLD) and make a series of subtle
adjustments to interest rates has elicited reactions from the financial market,
particularly in emerging economies that rely on the economic condition of the
Eurozone. These changes, aimed at stabilizing the Euro and promoting economic
growth in the European Union, have consequences far beyond the Eurozone region,
including capital flight, currency fluctuations, and increased risk aversion
among investors. The ECB's foresight in providing clarity and predictability is
met with mixed reactions, as some market analysts believe it lacks the
flexibility to adapt to diverse and dynamic economic conditions in various
regions. In response to this, smaller financial institutions and central banks
in emerging economies must take a series of adequate measures, ranging from
reducing interest rates to diversifying investment portfolios, to protect their
economies from external shocks and to capitalize on emerging opportunities. The
interaction between these global and local economic strategies highlights the
complex and interconnected nature of modern financial systems, in which the
decisions of a major player can profoundly influence the world economy.

\xhdr{Node 1-0-2 (Level 3, NV-Embed-v2 similarity to root 0.8955)} In the constant scenario of global economic change, the recent measures of the
European Central Bank (ECB) stand out as a significant sign of emerging
directions in international finance. The ECB's decision to modulate its
Quantitative Easing (QE) programs and a series of subtle adjustments to interest
rates have had repercussions on financial markets, especially in emerging
countries closely linked to the economic health of the eurozone. These changes,
aimed at stabilizing the euro and promoting growth in the European Union, have
consequences beyond, including capital flight, currency volatility, and
increased risk aversion among investors. The guidance provided by the ECB for
the future, with the aim of providing clarity and predictability, is received
with mixed reactions, as some market analysts suggest that it lacks the
flexibility needed to adapt to dynamic and diverse economic conditions in
different regions. As a response, smaller financial institutions and central
banks of emerging countries are obliged to adopt a variety of adaptive measures,
from reducing interest rates to diversifying investment portfolios, in order to
protect their economies from external shocks and seize emerging opportunities.
The interaction between these global and local economic strategies highlights
the complex and interconnected nature of modern financial systems, where the
decisions of a major player can have profound impacts on the rest of the world.

\xhdr{Node 1-1 (Level 2, NV-Embed-v2 similarity to root 0.9165)} In the constantly changing landscape of global economic and political
conditions, the recent measures of the European Central Bank (ECB) stand out as
a significant indicator of developing trends in the international financial
world. The ECB's decision to adjust its Quantitative Easing (QE) programs and
implement a series of fine-tuned interest rate adjustments triggers shockwaves
in financial markets, especially in emerging economies closely linked to the
economic health of the Eurozone. These changes, aimed at stabilizing the euro
and promoting growth in the European Union, have far-reaching consequences,
including capital flight, currency optimality, and increased risk aversion among
investors. The forward-looking but cautious approach of the ECB, which provides
clarity and predictability, meets with mixed reactions, as some market analysts
argue that it lacks the necessary flexibility to accommodate the diverse and
dynamic economic conditions in various regions. In response, smaller financial
institutions and central banks in emerging economies are forced to take a range
of adaptive measures, from interest rate reticence to diversifying their
investment portfolios, in order to shield their economies from external shocks
and seize new opportunities. The interaction between these global and local
economic strategies highlights the complex and intertwined nature of modern
financial systems, where the decisions of one major player can have profound
impacts on the rest of the world.

\xhdr{Node 1-1-0 (Level 3, NV-Embed-v2 similarity to root 0.9070)} In the constantly evolving global economic and political landscape, the recent
measures of the European Central Bank (ECB) stand out as a significant indicator
of trends developing in the international financial world. The decision of the
ECB to modify its Quantitative Easing (QE) programs and apply a series of
delicate interest rate adjustments has repercussions in financial markets,
particularly in emerging economies closely linked to the economic health of the
eurozone. These changes, aimed at stabilizing the euro and fostering growth in
the European Union, have far-reaching consequences, including capital flight,
currency optimization, and increased investor aversion. The ECB's forward-
looking and prudent approach, which brings clarity and predictability, meets
with mixed reactions, as some market analysts claim it lacks sufficient
flexibility to adapt to the diverse and dynamic economic conditions of various
regions. In response, smaller financial institutions and central banks of
emerging economies are compelled to adopt a range of adaptive measures, from
interest rate restraint to portfolio diversification, to protect their economies
from external shocks and seize new opportunities. The interaction between these
global and local economic strategies highlights the complex and interwoven
nature of modern financial systems, where the decisions of a major player can
have profound impacts on the rest of the world.

\xhdr{Node 1-1-1 (Level 3, NV-Embed-v2 similarity to root 0.8573)} In the constantly changing landscape of global economic and political
conditions, the recent measures of the European Central Bank (ECB) serve as an
important indicator of developing trends in the international financial world.
The decision of the ECB to adjust its Quantitative Easing programs and implement
a series of nuanced interest rate adjustments has waves in the financial
markets, especially in development-assisted economies closely linked to the
economic health of the Eurozone. These changes, aimed at stabilizing the Euro
and promoting growth in the European Union, have far-reaching consequences,
including capital flight, currency optimality, and increased risk aversion among
investors. The ECB's advancing but cautious approach, which offers clarity and
predictability, garners mixed reactions, as some market analysts argue that it
lacks the necessary flexibility to take into account the diverse and dynamic
economic conditions in various regions. As a result, smaller financial
institutions and central banks in development-assisted economies must take a
series of adaptive measures, ranging from interest rate restraint to
diversification of their investment portfolios, to protect their economies from
external shocks and seize new opportunities. The interaction between these
global and local economic strategies highlights the complex and interwoven
nature of modern financial systems, where the decisions of a key player can have
strong impacts on the rest of the world.

\xhdr{Node 1-1-2 (Level 3, NV-Embed-v2 similarity to root 0.9194)} In the ever-changing landscape of the global economic and political scenario,
the recent measures of the European Central Bank (ECB) stand out as a
significant indicator of emerging trends in the international financial world.
The decision of the ECB to adjust its Quantitative Easing (QE) programs and
implement a series of refined interest rate adjustments triggers shockwaves in
financial markets, particularly in emerging economies closely linked to the
economic health of the Eurozone. These changes, aimed at stabilizing the euro
and fostering growth in the European Union, have far-reaching consequences,
including capital flight, currency optimization, and increased risk aversion
among investors. The forward-looking but cautious approach of the ECB, which
provides clarity and predictability, receives mixed reactions, as some market
analysts argue that it lacks the necessary flexibility to adapt to dynamically
and diversely changing economic conditions in various regions. As a response,
smaller financial institutions and central banks of emerging economies are
compelled to adopt a series of adaptive measures, ranging from reluctance to
adjust interest rates to diversifying their investment portfolios, in order to
protect their economies from external shocks and seize new opportunities. The
interaction between these global and local economic strategies highlights the
complex and intertwined nature of modern financial systems, where the decisions
of one key player can have a profound impact on the rest of the world.

\xhdr{Node 1-2 (Level 2, NV-Embed-v2 similarity to root 0.9204)} In the constantly evolving global landscape of economic policy, the recent
measures of the European Central Bank (ECB) stand out as a significant sign of
evolving directions in the international financial world. The decision of the
ECB to adjust its quantitative easing (QE) programs and a series of subtle
interest rate adjustments have sparked shocks in financial markets, especially
in emerging economies closely tied to the economic health of the eurozone. These
changes, aimed at stabilizing the euro and promoting growth in the European
Union, have far-reaching consequences, including capital flight, currency
volatility, and increased risk aversion among investors. The future orientation
of the ECB, which seeks to provide clarity and predictability, is met with mixed
reactions, as some market analysts argue that it lacks the necessary flexibility
to address the diverse and dynamic economic conditions in different regions. As
a response, small financial institutions and central banks in emerging economies
are compelled to take a range of adaptive measures, from tightening interest
rate conditions to diversifying their investment portfolios, to protect their
economies from external shocks and seize emerging opportunities. The interaction
between these global and local economic strategies highlights the complex and
interconnected nature of modern financial systems, where decisions by a major
player can have profound impacts on the rest of the world.

\xhdr{Node 1-2-0 (Level 3, NV-Embed-v2 similarity to root 0.8980)} In the constantly evolving global economic landscape, the recent measures of the
European Central Bank (ECB) stand out as a significant sign of emerging
directions in the international financial world. The ECB's decision to modify
its quantitative easing (QE) programs and a series of subtle adjustments to
interest rates have caused ripples in financial markets, particularly in
emerging economies closely linked to the economic health of the eurozone. These
changes, aimed at stabilizing the euro and promoting development within the
European Union, have far-reaching consequences, including capital flight,
currency volatility, and increased risk aversion among investors. The future
orientation of the ECB, which seeks to bring clarity and predictability, is met
with mixed reactions, as some market analysts believe it lacks the necessary
flexibility to respond to diverse and dynamic economic conditions in different
regions. In response, financial institutions and central banks in emerging
economies are forced to take a range of adaptive measures, from tightening
interest rate conditions to diversifying their investment portfolios, to protect
their economies against external shocks and seize emerging opportunities. The
interaction between these global and local economic strategies highlights the
complex and interconnected nature of modern financial systems, where the
decisions of a major actor can have deep impacts on the rest of the world.

\xhdr{Node 1-2-1 (Level 3, NV-Embed-v2 similarity to root 0.9313)} In the constantly changing global landscape of international finance and
economic policy, the recent measures of the European Central Bank (ECB) serve as
an important indicator of evolving developments. The ECB's decision to adjust
its Quantitative Easing (QE) programs and implement a series of subtle interest
rate adjustments has sparked shocks in financial markets, particularly in
emerging economies that have close ties to the economic health of the Eurozone.
These changes, aimed at stabilizing the Euro and fostering economic growth in
the European Union, have far-reaching consequences, including capital flight,
currency fluctuations, and increased risk aversion among investors. The forward-
looking orientation of the ECB, which seeks to provide clear and predictable
measures, is met with mixed reactions, as some market analysts argue that it
lacks the necessary flexibility to address the diverse and dynamic economic
conditions in different regions. In response, small financial institutions and
central banks in emerging economies are obligated to take a range of adaptive
measures, from revealing interest rates to diversifying their investment
portfolios, to protect their economies from external shocks and seize new
opportunities. The interaction between these global and local economic
strategies highlights the complex and intertwined nature of modern financial
systems, where the decisions of a major player can have profound impacts on the
rest of the world.

\xhdr{Node 1-2-2 (Level 3, NV-Embed-v2 similarity to root 0.9081)} In the constantly evolving global economic landscape, recent measures by the
European Central Bank (ECB) stand out as a significant sign of changing
directions in the international financial world. The ECB's decision to adjust
its quantitative easing (QE) programs and a series of subtle adjustments in
interest rates have caused shocks in financial markets, especially in emerging
countries closely linked to the economic health of the eurozone. These changes,
aimed at stabilizing the euro and promoting growth in the European Union, have
far-reaching consequences, including capital flight, currency volatility, and
increased risk aversion among investors. The ECB's future orientation, seeking
to provide clarity and predictability, is received with mixed reactions, as some
market analysts argue that it lacks the necessary flexibility to address diverse
and dynamic economic conditions in different regions. As a response, small
financial institutions and central banks of emerging countries are compelled to
take a series of adaptive measures, from tightening interest rate conditions to
diversifying their investment portfolios, to protect their economies from
external shocks and capitalize on emerging opportunities. The interaction
between these global and local economic strategies highlights the complex and
interconnected nature of modern financial systems, where the decisions of one
major player can have a profound impact on the rest of the world.

\xhdr{Node 2 (Level 1, NV-Embed-v2 similarity to root 0.9469)} In the constantly evolving landscape of global economic governance, the recent
political adjustments by the European Central Bank (ECB) stand out as a
significant marker of changes in international financial waters. The ECB's
decision to recalibrate its quantitative easing (QE) program, along with a
series of subtle interest rate adjustments, has sent shockwaves through
financial markets, particularly affecting emerging countries closely linked to
the economic health of the Eurozone. These changes, designed to stabilize the
euro and stimulate economic growth within the European Union, have had broader
implications, including capital flight, currency volatility, and an increase in
risk aversion among investors. The ECB's forward guidance, intended to provide
clarity and predictability, has received mixed reactions, as some market
analysts argue that it lacks sufficient flexibility to address the diverse and
dynamic economic conditions in different regions. In response, smaller financial
institutions and central banks in emerging countries have been forced to adopt a
range of adaptive measures, from tightening monetary policies to diversifying
their investment portfolios, in order to insulate themselves from external
shocks and capitalize on emerging opportunities. The interplay between these
global and local economic strategies underscores the complex and interconnected
nature of the modern financial system, where the decisions of one major player
can have profound implications for the rest of the world.

\xhdr{Node 2-0 (Level 2, NV-Embed-v2 similarity to root 0.9227)} In the constantly evolving global economic landscape, the recent policy
adjustments of the European Central Bank (ECB) stand out as a significant marker
of changes in international financial waters. The ECB's decision to recalibrate
its quantitative easing (QE) program, along with a series of subtle adjustments
to interest rates, has caused ripples in financial markets, particularly
affecting emerging countries closely linked to the economic health of the
eurozone. These changes, designed to stabilize the euro and stimulate economic
growth within the European Union, have broader implications, including capital
flight, currency volatility, and increased risk aversion among investors. The
forward guidance, aimed at providing clarity and predictability, has received
mixed reactions, as some market analysts believe it lacks the necessary
flexibility to address the complex and changing economic conditions in different
regions. In response, smaller financial institutions and central banks in
emerging countries have had to take a series of adaptive measures, from
tightening monetary policy to diversifying portfolios, to protect themselves
from external shocks and seize emerging opportunities. The interaction between
global and regional economic strategies highlights the complexity and
interconnectivity of modern financial systems, where the decisions of one major
player can have far-reaching effects on the rest of the world.

\xhdr{Node 2-0-0 (Level 3, NV-Embed-v2 similarity to root 0.9002)} In the constantly evolving global economic landscape, the recent policy
adjustments by the European Central Bank (ECB) stand out as a significant sign
of changes on the horizon for international financial waters. The ECB's decision
to recalibrate its quantitative easing (QE) program, along with a series of
subtle interest rate adjustments, has caused ripples in financial markets,
particularly affecting emerging economies closely linked to the economic health
of the eurozone. These changes, designed to stabilize the euro and stimulate
economic growth within the European Union, have broader implications, including
capital flight, currency volatility, and increased investor aversion. The future
direction, aimed at bringing clarity and predictability, has received mixed
reactions, as some market analysts believe it lacks the necessary flexibility to
respond to complex and changing economic conditions in different regions. In
response, small financial institutions and central banks of emerging economies
have had to take a series of adaptive measures, ranging from tightening monetary
policy to diversifying portfolios, to protect themselves from external shocks
and seize emerging opportunities. The interaction between global and regional
economic strategies highlights the complexity and interconnectedness of modern
financial systems, where the decisions of a major actor can have reverberating
effects on the rest of the world.

\xhdr{Node 2-0-1 (Level 3, NV-Embed-v2 similarity to root 0.9184)} In the constantly changing global economic landscape, the recent policy changes
of the European Central Bank (ECB) are a significant marker for changes in
international financial waters. The ECB's decision to recalibrate its
Quantitative Easing (QE) program, along with a series of subtle adjustments to
interest rates, has sent waves through the financial markets, particularly in
Emerging Markets that are closely linked to the economic well-being of the
Eurozone. These changes, aimed at stabilizing the Euro and promoting economic
development within the European Union, have broader implications, including
capital flight, currency volatility, and increased risk aversion among
investors. The forward guidance strategy, which seeks to provide clarity and
predictability, is met with mixed reactions as some market analysts believe it
lacks the necessary flexibility to be fair to complex and changing economic
conditions in various regions. In response, smaller financial institutions and
central banks in Emerging Markets have taken a range of adaptive measures, from
abstaining from monetary policy to diversifying portfolios, to protect
themselves from external shocks and seize new opportunities. The interaction
between global and regional economic strategies highlights the complexity and
interdependence of modern financial systems, where the decisions of a major
player can have far-reaching effects on the rest of the world.

\xhdr{Node 2-0-2 (Level 3, NV-Embed-v2 similarity to root 0.8992)} In the constant scenario of the evolving global economic landscape, the recent
political adjustments by the European Central Bank (ECB) stand out as a
significant marker of changes in international financial waters. The decision of
the ECB to recalibrate its quantitative easing (QE) program, along with a series
of subtle adjustments in interest rates, has caused ripples in financial
markets, particularly affecting emerging countries that are closely linked to
the economic health of the eurozone. These changes, designed to stabilize the
euro and stimulate economic growth within the European Union, have broader
implications, including capital flight, exchange rate volatility, and increased
risk aversion among investors. The forward-looking orientation, aimed at
providing clarity and predictability, has received mixed reactions, as some
market analysts believe it lacks the necessary flexibility to address changing
and complex economic conditions in different regions. As a response, smaller
financial institutions and central banks of emerging countries have had to take
a series of adaptive measures, ranging from tightening monetary policy to
diversifying portfolios, to protect themselves from external shocks and seize
emerging opportunities. The interaction between global and regional economic
strategies highlights the complexity and interconnectivity of modern financial
systems, where the decisions of one major player can have far-reaching effects
on the rest of the world.

\xhdr{Node 2-1 (Level 2, NV-Embed-v2 similarity to root 0.9394)} In the constantly evolving landscape of global economic governance, the most
recent political adjustment of the European Central Bank (ECB) stands as an
important landmark for changes in the international financial waters. The
decision of the ECB to recalibrate its Quantitative Easing (QE) program, as well
as a series of slight interest rate adjustments, have unleashed shockwaves in
the financial markets, particularly in emerging countries with close ties to the
economic health of the Eurozone. These changes, aimed at stabilizing the Euro
and strengthening economic growth within the European Union, have far-reaching
implications, including capital flight, currency volatility, and an increase in
risk aversion among investors. The ECB's forecasts, which are intended to
provide clarity and predictability, have received mixed reactions, as some
market analysts argue that they do not offer enough flexibility to account for
the diverse and dynamic economic conditions in various regions. In response,
smaller financial institutions and central banks in emerging countries have
taken a range of adaptive measures, from lowering monetary policies to
diversifying their investment portfolios, to insulate themselves from external
shocks and take advantage of new opportunities. The interaction between these
global and local economic strategies underscores the complex and intertwined
nature of the modern financial system, where the decisions of one major player
can have significant implications for the rest of the world.

\xhdr{Node 2-1-0 (Level 3, NV-Embed-v2 similarity to root 0.9184)} In the constantly evolving global economic landscape, the recent adjustment of
the monetary policy of the European Central Bank (ECB) serves as an important
landmark for changes in international financial waters. The decision of the ECB
to recalibrate its Quantitative Easing (QE) program, along with a series of
minor adjustments to interest rates, has caused ripples in financial markets,
particularly among emerging economies closely linked to the economic health of
the eurozone. These changes, aimed at stabilizing the euro and strengthening
economic growth in the European Union, have wide-ranging implications, including
capital flight, currency volatility, and an increased risk appetite among
investors. The ECB's forecasts, which aim to bring clarity and predictability,
have received mixed reactions, as some market analysts believe they do not
provide enough flexibility to account for diverse and dynamic economic
conditions in different regions. In response, financial institutions and central
banks of emerging economies have adopted a range of adaptive measures, from
easing monetary policies to diversifying investment portfolios, to protect
against external shocks and take advantage of new opportunities. The interaction
between these global and local economic strategies highlights the complex and
interconnected nature of the modern financial system, where the decisions of a
major player can have significant implications for the rest of the world.

\xhdr{Node 2-1-1 (Level 3, NV-Embed-v2 similarity to root 0.8603)} In the constantly changing landscape of global economic and political
governance, the latest political change of the European Central Bank (ECB)
represents an important milestone for changes in international financial waters.
The decision of the ECB to recalibrate its Quantitative Easing (QE) program, as
well as a series of minor interest rate adjustments, has triggered waves of
shockwaves in the financial market, particularly in development countries that
rely on the economic health of the Eurozone. These changes, aimed at stabilizing
the Euro and strengthening economic growth in the European Union, have far-
reaching implications, including capital flight, currency volatility, and an
increase in risk aversion among investors. The ECB's forecasts, which aim to
increase clarity and predictability, have received mixed reactions, as some
market analysts argue that they are not sufficiently flexible to accommodate the
diverse and dynamic economic conditions in different regions. In response,
smaller financial institutions and central banks in developing countries have
taken a series of adaptable measures, ranging from reducing monetary policy to
diversifying their investment portfolios, to insulate themselves from external
shocks and seize new opportunities. The interaction between these global and
local economic strategies underscores the complex and interwoven nature of the
modern financial system, where the decisions of one major player can have
significant impacts on the rest of the world.

\xhdr{Node 2-1-2 (Level 3, NV-Embed-v2 similarity to root 0.9302)} In the constant scenario of the evolution of global economic governance, the
latest political adjustment by the European Central Bank (ECB) has risen to
become an important milestone for changes in international financial waters. The
ECB's decision to recalibrate its Quantitative Easing (QE) program, along with a
series of minor adjustments to interest rates, has triggered shockwaves in
financial markets, especially in emerging economies with close ties to the
economic health of the Eurozone. These changes, aimed at stabilizing the Euro
and strengthening economic growth within the European Union, have long-term
implications, including capital flight, currency volatility, and an increase in
risk aversion among investors. The ECB's projections, which seek to provide
clarity and predictability, have received mixed reactions, as some market
analysts argue that they do not offer sufficient flexibility to address the
diverse and dynamic economic conditions in various regions. In response,
financial institutions and central banks of emerging economies have taken a
series of adaptive measures, from lowering their monetary policies to
diversifying their investment portfolios, to protect themselves from external
shocks and seize new opportunities. The interaction between these global and
local economic strategies highlights the complex and interconnected nature of
the modern financial system, where decisions by a single dominant player can
have significant implications for the rest of the world.

\xhdr{Node 2-2 (Level 2, NV-Embed-v2 similarity to root 0.9398)} In the evolving landscape of global economic governance, the recent political
adjustments by the European Central Bank (ECB) stand out as a significant marker
of changes in international financial waters. The ECB's decision to recalibrate
its quantitative easing (QE) program, along with a series of subtle adjustments
in interest rates, has sent shockwaves through financial markets, particularly
affecting emerging countries closely linked to the economic health of the
European Economic Area. These changes, designed to stabilize the euro and
stimulate economic growth within the European Union, have broader implications,
including capital flight, currency volatility, and an increase in risk aversion
among investors. The future guidance provided by the ECB, intended to offer
clarity and predictability, has received mixed reactions, as some market
analysts argue that it lacks sufficient flexibility to address the dynamic and
diverse economic conditions in different regions. In response, smaller financial
institutions and central banks of emerging countries have been forced to adopt a
range of adaptive measures, from tightening monetary policies to diversifying
their investment portfolios, with the aim of insulating themselves from external
shocks and capitalizing on emerging opportunities. The interaction between these
global and local economic strategies highlights the complex and interconnected
nature of the modern financial system, where the decisions of a single major
player can have profound implications for the rest of the world.

\xhdr{Node 2-2-0 (Level 3, NV-Embed-v2 similarity to root 0.9372)} In the constantly evolving landscape of global economic governance, the recent
political adjustments by the European Central Bank (ECB) stand out as a
significant marker of changes in international financial waters. The ECB's
decision to recalibrate its quantitative easing (QE) program, along with a
series of subtle adjustments to interest rates, sent shockwaves through
financial markets, particularly affecting emerging economies closely tied to the
economic health of the European Economic Area. These changes, designed to
stabilize the euro and stimulate economic growth in the European Union, have
broader implications, including currency volatility and an increase in risk
aversion among investors. The future guidance provided by the ECB, aimed at
offering clarity and predictability, has received mixed reactions, as some
financial analysts believe it lacks flexibility to respond to dynamic and
diverse economic conditions in different regions. In response, smaller financial
institutions and central banks of emerging economies have been forced to adopt a
range of adaptive measures, from tightening monetary policies to diversifying
their investment portfolios, with the goal of insulating these countries from
external shocks and capitalizing on emerging opportunities. The interaction
between these global and local economic strategies highlights the complex and
interconnected nature of the modern financial system, where the decisions of a
single major actor can have profound implications for the rest of the world.

\xhdr{Node 2-2-1 (Level 3, NV-Embed-v2 similarity to root 0.9168)} In the constantly evolving scenario of global economic and political governance,
the recent political adjustment by the European Central Bank (ECB) represents a
significant feature of changes in the international financial waters. The ECB's
decision to recalibrate its Quantitative Easing (QE) program, as well as a
series of subtle adjustments to interest rates, have triggered shockwaves in the
financial markets, particularly in emerging countries closely linked to the
economic health of the European Economic and Monetary Union (EMU). These
changes, aimed at stabilizing the euro and promoting economic development in the
European Union, have broader impacts, including capital flight, currency
instability, and increased risk aversion among investors. The future guidance
provided to the ECB to offer clarity and predictability has elicited mixed
reactions, as some market analysts argue that it does not offer sufficient
flexibility to account for the dynamic and diverse economic conditions in
various regions. In response, smaller financial institutions and central banks
in emerging countries have taken a range of adaptive measures, from the
abstinence of monetary policy to the diversification of their investor
portfolios, with the goal of insulating themselves from external shocks and
capitalizing on emerging opportunities. The interaction between these global and
local economic strategies highlights the complex and interconnected nature of
the modern financial system, where decisions of a single major player can have
far-reaching impacts on the rest of the world.

\xhdr{Node 2-2-2 (Level 3, NV-Embed-v2 similarity to root 0.9388)} In the evolutionary landscape of global economic governance, the recent
political adjustments made by the European Central Bank (ECB) stand out as a
significant marker of changes in international financial waters. The ECB's
decision to recalibrate its quantitative easing (QE) program, along with a
series of subtle adjustments in interest rates, has sent shockwaves through
financial markets, particularly affecting emerging countries that are closely
linked to the economic health of the Euro Area. These changes, designed to
stabilize the euro and stimulate economic growth within the European Union, have
broader implications, including capital flight, currency volatility, and
increased risk aversion among investors. The guidance provided by the ECB,
intended to offer clarity and predictability, has received mixed reactions, as
some market analysts argue that it lacks sufficient flexibility to address the
dynamic and diverse economic conditions in different regions. In response,
smaller financial institutions and central banks of emerging countries have been
forced to adopt a series of adaptive measures, ranging from tightening monetary
policies to diversifying their investment portfolios, with the goal of
insulating themselves from external shocks and capitalizing on emerging
opportunities. The interaction between these global and local economic
strategies illustrates the complex and interconnected nature of the modern
financial system, where the decisions of a single major player can have profound
implications for the rest of the world.
}
\\
This is the full content of the self-consistency tree generated by Qwen-2.5-72B on machine translation tasks.

{\ttfamily\small
\xhdr{Root (Level 0, NV-Embed-v2 similarity to root 1.0000)} In the ever-evolving landscape of global economic governance, the recent policy
adjustments by the European Central Bank (ECB) stand out as a significant marker
of the shifting tides in international finance. The ECB's decision to
recalibrate its quantitative easing (QE) program, alongside a series of nuanced
interest rate adjustments, has sent shockwaves through the financial markets,
particularly affecting emerging economies that are intricately linked to the
Eurozone's economic health. These changes, while designed to stabilize the Euro
and stimulate economic growth within the European Union, have had far-reaching
consequences, including capital flight, currency volatility, and heightened risk
aversion among investors. The ECB's forward guidance, which aims to provide
clarity and predictability, has been met with mixed reactions, as some market
analysts argue that it lacks the necessary flexibility to address the diverse
and dynamic economic conditions across different regions. In response, smaller
financial institutions and central banks in emerging markets have been compelled
to adopt a range of adaptive measures, from tightening monetary policies to
diversifying their investment portfolios, in an effort to insulate their
economies from external shocks and capitalize on the emerging opportunities. The
interplay between these global and local economic strategies underscores the
complex and interconnected nature of the modern financial system, where the
decisions of one major player can have profound implications for the rest of the
world.

\xhdr{Node 0 (Level 1, NV-Embed-v2 similarity to root 0.9745)} In the constantly evolving landscape of global economic governance, recent
policy adjustments by the European Central Bank (ECB) stand out as a significant
marker of trends in international finance. The ECB's decision to recalibrate its
quantitative easing (QE) program, coupled with a series of subtle interest rate
adjustments, has sent shockwaves through financial markets, particularly
affecting emerging economies that are closely tied to the economic health of the
eurozone. These changes, although designed to stabilize the euro and stimulate
economic growth within the European Union, have had long-term consequences,
including capital flight, currency volatility, and increased risk aversion among
investors. The ECB's forward guidance, aimed at providing clarity and
predictability, has been met with mixed reactions, some market analysts arguing
that it lacks the necessary flexibility to respond to the diverse and dynamic
economic conditions across different regions. In response, smaller financial
institutions and central banks in emerging markets have been forced to adopt a
range of adaptive measures, from tightening monetary policies to diversifying
their investment portfolios, in an effort to protect their economies from
external shocks and capitalize on emerging opportunities. The interaction
between these global and local economic strategies underscores the complex and
interconnected nature of the modern financial system, where decisions by a major
player can have profound implications for the rest of the world.

\xhdr{Node 0-0 (Level 2, NV-Embed-v2 similarity to root 0.9672)} In the constantly evolving landscape of global economic governance, the recent
policy adjustments by the European Central Bank (ECB) stand out as a significant
marker of trends in international finance. The ECB's decision to recalibrate its
quantitative easing (QE) program, coupled with a series of subtle interest rate
adjustments, has sent shockwaves through financial markets, particularly
impacting emerging economies closely tied to the economic health of the
eurozone. Although these changes are designed to stabilize the euro and boost
economic growth within the European Union, they have had long-term consequences,
including capital flight, currency volatility, and increased risk aversion among
investors. Future guidance from the ECB, aimed at providing clarity and
predictability, has been met with mixed reactions, some market analysts arguing
that it lacks the necessary flexibility to respond to the diverse and dynamic
economic conditions across different regions. In response, smaller financial
institutions and central banks in emerging markets have been forced to adopt a
range of adaptive measures, from tightening monetary policies to diversifying
their investment portfolios, in an effort to protect their economies from
external shocks and capitalize on emerging opportunities. The interaction
between these global and local economic strategies underscores the complex and
interconnected nature of the modern financial system, where decisions by a major
player can have profound implications for the rest of the world.

\xhdr{Node 0-0-0 (Level 3, NV-Embed-v2 similarity to root 0.9672)} In the constantly evolving landscape of global economic governance, the recent
policy adjustments by the European Central Bank (ECB) stand out as a significant
marker of trends in international finance. The ECB's decision to recalibrate its
quantitative easing (QE) program, coupled with a series of subtle interest rate
adjustments, has sent shockwaves through financial markets, particularly
affecting emerging economies closely tied to the economic health of the
eurozone. Although these changes are designed to stabilize the euro and
stimulate economic growth within the European Union, they have had long-term
consequences, including capital flight, currency volatility, and increased risk
aversion among investors. The ECB's future guidance, aimed at providing clarity
and predictability, has been met with mixed reactions, some market analysts
arguing that it lacks the necessary flexibility to respond to the diverse and
dynamic economic conditions across different regions. In response, smaller
financial institutions and central banks in emerging markets have been forced to
adopt a range of adaptation measures, from tightening monetary policies to
diversifying their investment portfolios, in an effort to protect their
economies from external shocks and capitalize on emerging opportunities. The
interaction between these global and local economic strategies underscores the
complex and interconnected nature of the modern financial system, where
decisions by a major player can have profound implications for the rest of the
world.

\xhdr{Node 0-0-1 (Level 3, NV-Embed-v2 similarity to root 0.9424)} In the constantly changing landscape of global economic governance, the recent
political adjustments by the European Central Bank (ECB) are a significant
indicator of trends in the international financial world. The ECB's decision to
recalibrate its Quantitative Easing (QE) program, coupled with a series of
subtle interest rate adjustments, has sent shockwaves through financial markets,
particularly to emerging economies that are closely tied to the economic health
of the Eurozone. While these changes are designed to stabilize the Euro and
promote economic growth within the European Union, they have had long-term
consequences, including capital flight, currency fluctuations, and increased
risk aversion among investors. The future guideline of the ECB, which aims to
create clarity and predictability, has been met with mixed reactions, with some
market analysts arguing that it lacks the necessary flexibility to respond to
the diverse and dynamic economic conditions in different regions. In response,
smaller financial institutions and central banks in emerging markets have been
forced to adopt a range of adaptive measures, from tightening monetary policy to
diversifying their investment portfolios, to protect their economies from
external shocks and capitalize on new opportunities. The interaction between
these global and local economic strategies underscores the complex and
interconnected nature of the modern financial system, where decisions by a key
player can have far-reaching impacts on the rest of the world.

\xhdr{Node 0-0-2 (Level 3, NV-Embed-v2 similarity to root 0.9677)} In the constantly evolving landscape of global economic governance, recent
policy adjustments by the European Central Bank (ECB) stand out as a significant
marker of trends in international finance. The ECB's decision to recalibrate its
quantitative easing (QE) program, along with a series of subtle interest rate
adjustments, has sent shockwaves through financial markets, particularly
impacting emerging economies closely tied to the economic health of the
eurozone. While these changes are designed to stabilize the euro and stimulate
economic growth within the European Union, they have had long-term consequences,
including capital flight, exchange rate volatility, and an increase in risk
aversion among investors. The ECB's future guidance, aimed at providing clarity
and predictability, has been met with mixed reactions, some market analysts
arguing that it lacks the necessary flexibility to respond to the diverse and
dynamic economic conditions in different regions. In response, financial
institutions and smaller central banks in emerging markets have been forced to
adopt a series of adaptive measures, from tightening monetary policies to
diversifying their investment portfolios, in an effort to protect their
economies from external shocks and capitalize on emerging opportunities. The
interaction between these global and local economic strategies underscores the
complex and interconnected nature of the modern financial system, where
decisions by a major player can have profound implications for the rest of the
world.

\xhdr{Node 0-1 (Level 2, NV-Embed-v2 similarity to root 0.9274)} In the constantly changing landscape of global economic governance, the recent
political adjustments by the European Central Bank (ECB) are a significant
indicator of trends in the international financial world. The ECB's decision to
recalibrate its Quantitative Easing (QE) program, coupled with a series of
subtle interest rate changes, has sent shockwaves through financial markets,
particularly to emerging countries closely tied to the economic health of the
Eurozone. While these changes aim to stabilize the Euro and promote economic
growth within the European Union, they have long-term consequences, including
capital flight, currency volatility, and increased risk aversion among
investors. The future leadership of the ECB, which is intended to provide
clarity and predictability, has been met with mixed reactions, with some market
analysts arguing that it lacks the necessary flexibility to respond to the
diverse and dynamic economic conditions in different regions. In response,
smaller financial institutions and central banks in emerging markets have been
forced to adopt a range of adaptive measures, from tightening monetary policies
to diversifying their investment portfolios, to protect their economies from
external shocks and capitalize on opportunities. The interaction between these
global and local economic strategies underscores the complex and interconnected
nature of the modern financial system, where decisions by a major player can
have far-reaching effects on the rest of the world.

\xhdr{Node 0-1-0 (Level 3, NV-Embed-v2 similarity to root 0.9488)} In the constantly evolving landscape of global economic governance, the recent
policy adjustments by the European Central Bank (ECB) are a significant
indicator of trends in the international financial world. The ECB's decision to
adjust its Quantitative Easing (QE) program, along with a series of small
interest rate changes, has sent shockwaves through financial markets,
particularly to emerging countries closely tied to the economic health of the
eurozone. While these changes aim to stabilize the euro and promote economic
growth within the European Union, they have long-term consequences, including
capital outflows, exchange rate fluctuations, and increased risk aversion among
investors. The future direction of the ECB, which is supposed to provide clarity
and predictability, has been met with mixed reactions, some market analysts
arguing that it lacks the necessary flexibility to respond to the diverse and
dynamic economic conditions in different regions. In response, smaller financial
institutions and central banks in emerging markets have been forced to adopt a
range of adaptive measures, from tightening monetary policies to diversifying
their investment portfolios, to protect their economies from external shocks and
capitalize on opportunities. The interaction between these global and local
economic strategies underscores the complex and interconnected nature of the
modern financial system, where decisions by a major player can have long-term
effects on the rest of the world.

\xhdr{Node 0-1-1 (Level 3, NV-Embed-v2 similarity to root 0.9263)} In the constantly changing landscape of global economic governance, the recent
political adjustments by the European Central Bank (ECB) are a significant
indicator of trends in the international financial world. The ECB's decision to
recalibrate its Quantitative Easing (QE) program, coupled with a series of
subtle interest rate changes, has sent shockwaves through financial markets,
particularly in emerging countries that are closely tied to the economic health
of the Eurozone. While these changes aim to stabilize the euro and promote
economic growth within the European Union, they have long-term consequences,
including capital flight, currency volatility, and increased risk aversion among
investors. The future leadership of the ECB, which is expected to provide
clarity and predictability, has been met with mixed reactions, with some market
analysts arguing that it lacks the necessary flexibility to respond to the
diverse and dynamic economic conditions in different regions. In response,
smaller financial institutions and central banks in emerging markets have been
forced to implement a range of adjustment measures, from tightening monetary
policy to diversifying their investment portfolios, to protect their economies
from external shocks and capitalize on opportunities. The interaction between
these global and local economic strategies underscores the complex and
interconnected nature of the modern financial system, where decisions by a major
player can have far-reaching impacts on the rest of the world.

\xhdr{Node 0-1-2 (Level 3, NV-Embed-v2 similarity to root 0.9382)} In the constantly changing landscape of global economic governance, the recent
policy adjustments by the European Central Bank (ECB) are a significant
indicator of trends in international financial markets. The ECB's decision to
recalibrate its Quantitative Easing (QE) program, along with a series of subtle
changes in interest rates, has sent shockwaves through financial markets,
particularly to emerging countries closely tied to the economic health of the
Eurozone. While these changes aim to stabilize the Euro and promote economic
growth within the European Union, they have long-term consequences, including
capital flight, currency volatility, and an increase in risk aversion among
investors. The future leadership of the ECB, which is intended to provide
clarity and predictability, has been met with mixed reactions, with some market
analysts arguing that it lacks the necessary flexibility to respond to diverse
and dynamic economic conditions in different regions. In response, financial
institutions and smaller central banks in emerging markets have been forced to
adopt a range of adaptive measures, from tightening monetary policies to
diversifying their investment portfolios, to protect their economies from
external shocks and capitalize on opportunities. The interaction between these
global and local economic strategies underscores the complex and interconnected
nature of the modern financial system, where decisions by a major player can
have far-reaching effects on the rest of the world.

\xhdr{Node 0-2 (Level 2, NV-Embed-v2 similarity to root 0.9748)} In the constantly evolving landscape of global economic governance, recent
policy adjustments by the European Central Bank (ECB) stand out as a significant
marker of trends in international finance. The ECB's decision to recalibrate its
quantitative easing (QE) program, along with a series of subtle interest rate
adjustments, has sent shockwaves through financial markets, particularly
affecting emerging economies that are closely tied to the economic health of the
eurozone. These changes, although designed to stabilize the euro and stimulate
economic growth within the European Union, have had long-term consequences,
including capital flight, currency volatility, and an increase in risk aversion
among investors. The ECB's forward guidance, intended to provide clarity and
predictability, has been met with mixed reactions, some market analysts arguing
that it lacks the necessary flexibility to respond to the diverse and dynamic
economic conditions in different regions. In response, financial institutions
and smaller central banks in emerging markets have been forced to adopt a
variety of adaptive measures, from tightening monetary policies to diversifying
their investment portfolios, in an effort to protect their economies from
external shocks and capitalize on emerging opportunities. The interaction
between these global and local economic strategies underscores the complex and
interconnected nature of the modern financial system, where decisions by a major
player can have profound implications for the rest of the world.

\xhdr{Node 0-2-0 (Level 3, NV-Embed-v2 similarity to root 0.9653)} In the constantly evolving landscape of global economic governance, the recent
policy adjustments by the European Central Bank (ECB) stand out as a significant
marker of trends in international finance. The ECB's decision to recalibrate its
quantitative easing (QE) program, coupled with a series of subtle interest rate
adjustments, has sent shockwaves through financial markets, particularly
affecting emerging economies closely tied to the economic health of the
eurozone. These changes, although designed to stabilize the euro and boost
economic growth within the European Union, have had long-term consequences,
including capital outflows, exchange rate volatility, and increased risk
aversion among investors. The ECB's future guidance, aimed at providing clarity
and predictability, has been met with mixed reactions, some market analysts
arguing that it lacks the necessary flexibility to respond to the diverse and
dynamic economic conditions in different regions. In response, financial
institutions and central banks of emerging markets have been forced to adopt a
variety of adaptive measures, ranging from tightening monetary policies to
diversifying their investment portfolios, in an effort to protect their
economies from external shocks and capitalize on emerging opportunities. The
interaction between these global and local economic strategies underscores the
complex and interconnected nature of the modern financial system, where
decisions by a major player can have profound implications for the rest of the
world.

\xhdr{Node 0-2-1 (Level 3, NV-Embed-v2 similarity to root 0.9697)} In the constantly changing landscape of global economic governance, the recent
policy adjustments by the European Central Bank (ECB) stand out as a significant
indicator of trends in international finance. The ECB's decision to recalibrate
its quantitative easing (QE) program, along with a series of subtle interest
rate adjustments, has sent shockwaves through financial markets, particularly in
emerging countries that are closely tied to the economic health of the Eurozone.
These changes, although designed to stabilize the Euro and promote economic
growth within the European Union, had long-term consequences, including capital
flight, currency fluctuations, and an increase in investor risk aversion. The
ECB's forward guidance, intended to provide clarity and predictability, has been
met with mixed reactions. Some market analysts argue that it lacks the necessary
flexibility to respond to the diverse and dynamic economic conditions in
different regions. In response, financial institutions and smaller central banks
in emerging markets have been forced to adopt a range of adaptive measures, from
tightening monetary policies to diversifying their investment portfolios, to
protect their economies from external shocks and capitalize on new
opportunities. The interaction between these global and local economic
strategies underscores the complex and interconnected nature of the modern
financial system, where decisions by a major player can have far-reaching
effects on the rest of the world.

\xhdr{Node 0-2-2 (Level 3, NV-Embed-v2 similarity to root 0.9670)} In the constantly evolving landscape of global economic governance, recent
policy adjustments by the European Central Bank (ECB) stand out as a significant
marker of trends in international finance. The ECB's decision to recalibrate its
quantitative easing (QE) program, along with a series of subtle interest rate
adjustments, has sent ripples through financial markets, particularly affecting
emerging economies that are closely tied to the economic health of the eurozone.
These changes, although designed to stabilize the euro and stimulate economic
growth within the European Union, have had long-term consequences, including
capital flight, currency volatility, and an increase in risk aversion among
investors. The future guidance of the ECB, aimed at providing clarity and
predictability, has been met with mixed reactions, some market analysts arguing
that it lacks the necessary flexibility to respond to diverse and dynamic
economic conditions in different regions. In response, financial institutions
and smaller central banks in emerging markets have been forced to adopt a
variety of adaptive measures, from tightening monetary policies to diversifying
their investment portfolios, in an effort to protect their economies from
external shocks and capitalize on emerging opportunities. The interaction
between these global and local economic strategies underscores the complex and
interconnected nature of the modern financial system, where decisions by a major
player can have profound implications for the rest of the world.

\xhdr{Node 1 (Level 1, NV-Embed-v2 similarity to root 0.9453)} In the constantly changing landscape of global economic governance, the recent
political adjustments by the European Central Bank (ECB) are a significant sign
of the shifting currents in international finance. The ECB's decision to adjust
its Quantitative Easing (QE) program, along with a series of differentiated
interest rate adjustments, has sent shockwaves through the financial markets,
particularly affecting emerging economies that are closely tied to the economic
health of the Eurozone. These changes, aimed at stabilizing the Euro and
boosting economic growth in the European Union, have had far-reaching
consequences, including capital flight, currency fluctuations, and increased
risk aversion among investors. The forward-looking communication of the ECB,
intended to provide clarity and predictability, has been met with mixed
reactions, as some market analysts argue that it lacks the necessary flexibility
to accommodate the diverse and dynamic economic conditions in different regions.
In response, smaller financial institutions and central banks in developing
countries have been forced to implement a range of adaptive measures, from
tightening monetary policy to diversifying their investment portfolios, to
shield their economies from external shocks and capitalize on available
opportunities. The interaction between these global and local economic
strategies underscores the complex and interconnected nature of the modern
financial system, where decisions by a major player can have far-reaching
impacts on the rest of the world.

\xhdr{Node 1-0 (Level 2, NV-Embed-v2 similarity to root 0.9745)} In the constantly evolving landscape of global economic governance, the recent
policy adjustments by the European Central Bank (ECB) are a significant sign of
changing tides in international finance. The ECB's decision to adjust its
Quantitative Easing (QE) program, along with a series of differentiated interest
rate adjustments, has sent shockwaves through financial markets, particularly
affecting emerging economies that are closely tied to the economic health of the
eurozone. These changes, aimed at stabilizing the euro and stimulating economic
growth within the European Union, have had significant consequences, including
capital outflows, exchange rate fluctuations, and increased risk aversion among
investors. The forward guidance from the ECB, intended to provide clarity and
predictability, has been met with mixed reactions, some market analysts arguing
that it lacks the necessary flexibility to adapt to diverse and dynamic economic
conditions in different regions. In response, smaller financial institutions and
central banks of developing countries have been forced to implement a range of
adaptive measures, from tightening monetary policy to diversifying their
investment portfolios, to protect their economies from external shocks and
capitalize on available opportunities. The interaction between these global and
local economic strategies underscores the complex and interconnected nature of
the modern financial system, where decisions by a major player can have
significant impacts on the rest of the world.

\xhdr{Node 1-0-0 (Level 3, NV-Embed-v2 similarity to root 0.9586)} In the constantly evolving landscape of global economic governance, the recent
policy adjustments by the European Central Bank (ECB) are a significant sign of
changes in the currents of international finance. The ECB's decision to adjust
its Quantitative Easing (QE) program, along with a series of differentiated
interest rate adjustments, has sent shockwaves through financial markets,
particularly affecting emerging economies closely tied to the economic health of
the eurozone. These changes, aimed at stabilizing the euro and stimulating
economic growth within the European Union, have had significant consequences,
including capital outflows, exchange rate fluctuations, and increased risk
aversion among investors. The future guidance of the ECB, intended to bring
clarity and predictability, has been met with mixed reactions, some market
analysts arguing that they lack the necessary flexibility to adapt to the
diverse and dynamic economic conditions in different regions. In response, small
financial institutions and central banks of developing countries have been
forced to implement a range of adaptive measures, from tightening monetary
policy to diversifying their investment portfolios, to protect their economies
from external shocks and capitalize on available opportunities. The interaction
between these global and local economic strategies underscores the complex and
interconnected nature of the modern financial system, where decisions by a major
player can have significant impacts on the rest of the world.

\xhdr{Node 1-0-1 (Level 3, NV-Embed-v2 similarity to root 0.9468)} In the constantly changing environment of global economic governance, the recent
political adjustments by the European Central Bank (ECB) are a significant sign
of changes in the international financial world. The ECB's decision to adjust
its Quantitative Easing (QE) program, along with a series of differentiated
interest rate adjustments, has sent shockwaves through the financial markets,
especially the emerging economies that are closely tied to the economic health
of the Eurozone. These changes, aimed at stabilizing the euro and boosting
economic growth within the European Union, have had substantial consequences,
including capital outflows, exchange rate fluctuations, and increased risk
aversion among investors. The forward-looking guidance from the ECB, intended to
provide clarity and predictability, has been met with mixed reactions, with some
market analysts arguing that it lacks the necessary flexibility to adapt to the
diverse and dynamic economic conditions in different regions. In response,
smaller financial institutions and central banks of developing countries have
been forced to implement a range of adaptive measures, from tightening monetary
policy to diversifying their investment portfolios, to protect their economies
from external shocks and capitalize on existing opportunities. The interaction
between these global and local economic strategies underscores the complex and
interconnected nature of the modern financial system, where decisions by a key
player can have significant impacts on the rest of the world.

\xhdr{Node 1-0-2 (Level 3, NV-Embed-v2 similarity to root 0.9678)} In the constantly evolving landscape of global economic governance, recent
policy adjustments by the European Central Bank (ECB) are a significant sign of
changes in the tides of international finance. The ECB's decision to adjust its
Quantitative Easing (QE) program, along with a series of differentiated interest
rate adjustments, has sent shock waves through financial markets, particularly
affecting emerging economies that are closely tied to the economic health of the
eurozone. These changes, aimed at stabilizing the euro and stimulating economic
growth within the European Union, have had significant consequences, including
capital flows, exchange rate fluctuations, and an increase in risk aversion
among investors. The ECB's forward guidance, intended to provide clarity and
predictability, has been met with mixed reactions, with some market analysts
arguing that it lacks the necessary flexibility to adapt to diverse and dynamic
economic conditions in different regions. In response, smaller financial
institutions and central banks of developing countries have been forced to
implement a range of adaptive measures, from tightening monetary policy to
diversifying their investment portfolios, to protect their economies from
external shocks and take advantage of available opportunities. The interaction
between these global and local economic strategies underscores the complex and
interconnected nature of the modern financial system, where decisions by a major
player can have significant impacts on the rest of the world.

\xhdr{Node 1-1 (Level 2, NV-Embed-v2 similarity to root 0.9418)} In the constantly changing landscape of global economic governance, the recent
political adjustments by the European Central Bank (ECB) are a significant sign
of the shifting tides in the international financial world. The ECB's decision
to adjust its Quantitative Easing (QE) program, along with a series of different
interest rate adjustments, has sent shockwaves through the financial markets,
particularly the emerging economies that are closely tied to the economic
stability of the Eurozone. These changes, aimed at stabilizing the euro and
promoting economic growth in the European Union, have had far-reaching
consequences, including capital flight, currency fluctuations, and increased
risk aversion among investors. The forward-looking communication of the ECB,
which aims to provide clarity and predictability, has been met with mixed
reactions, as some market analysts argue that it lacks the necessary flexibility
to account for the diverse and dynamic economic conditions in different regions.
In response, smaller financial institutions and central banks in developing
countries have been forced to implement a range of adaptive measures, from
tightening monetary policy to diversifying their investment portfolios, to
protect their economies from external shocks and capitalize on available
opportunities. The interaction between these global and local economic
strategies underscores the complex and interconnected nature of the modern
financial system, where decisions by a key player can have far-reaching impacts
on the rest of the world.

\xhdr{Node 1-1-0 (Level 3, NV-Embed-v2 similarity to root 0.9603)} In the constantly evolving landscape of global economic governance, the recent
policy adjustments by the European Central Bank (ECB) are a significant sign of
trends changing in the international financial world. The ECB's decision to
adjust its Quantitative Easing (QE) program, along with a series of different
interest rate adjustments, has sent shockwaves through financial markets,
particularly emerging economies that are closely tied to the economic stability
of the eurozone. These changes, aimed at stabilizing the euro and promoting
economic growth within the European Union, have had significant consequences,
including capital outflows, exchange rate fluctuations, and increased risk
aversion among investors. The forward guidance from the ECB, which aims to
provide clarity and predictability, has received mixed reactions, with some
market analysts arguing that it lacks the necessary flexibility to account for
the diverse and dynamic economic conditions in different regions. In response,
smaller financial institutions and central banks of developing countries have
been forced to implement a range of adaptive measures, from tightening monetary
policy to diversifying their investment portfolios, to protect their economies
from external shocks and take advantage of available opportunities. The
interaction between these global and local economic strategies underscores the
complex and interconnected nature of the modern financial system, where
decisions by a key player can have significant impacts on the rest of the world.

\xhdr{Node 1-1-1 (Level 3, NV-Embed-v2 similarity to root 0.9420)} In the constantly changing landscape of global economic governance, the recent
political adjustments by the European Central Bank (ECB) are a significant sign
of the shifting currents in the international financial world. The ECB's
decision to adjust its Quantitative Easing (QE) program, along with a series of
interest rate changes, has sent shockwaves through financial markets,
particularly through emerging markets that are closely tied to the economic
stability of the Eurozone. These changes, aimed at stabilizing the euro and
promoting economic growth in the European Union, have had far-reaching
consequences, including capital flight, currency fluctuations, and increased
risk aversion among investors. The forward-looking communication of the ECB,
which aims to provide clarity and predictability, has been met with mixed
reactions, as some market analysts argue that it lacks the necessary flexibility
to accommodate the diverse and dynamic economic conditions in different regions.
In response, smaller financial institutions and central banks in developing
countries have been forced to implement a range of adaptive measures, from
tightening monetary policy to diversifying their investment portfolios, to
protect their economies from external shocks and capitalize on available
opportunities. The interaction between these global and local economic
strategies underscores the complex and interconnected nature of the modern
financial system, where decisions by a key player can have far-reaching impacts
on the rest of the world.

\xhdr{Node 1-1-2 (Level 3, NV-Embed-v2 similarity to root 0.9449)} In the constantly changing landscape of global economic governance, recent
political adjustments by the European Central Bank (ECB) are a significant sign
of changes in the international financial world. The ECB's decision to adjust
its Quantitative Easing (QE) program, along with a series of different interest
rate adjustments, has sent shockwaves through financial markets, particularly to
emerging economies that are closely tied to the economic stability of the
Eurozone. These changes, aimed at stabilizing the euro and promoting economic
growth in the European Union, have had far-reaching consequences, including
capital flight, currency fluctuations, and an increase in risk aversion among
investors. The forward guidance of the ECB, which aims to provide clarity and
predictability, has been met with mixed reactions, as some market analysts argue
that it lacks the necessary flexibility to account for the diverse and dynamic
economic conditions in different regions. In response, financial institutions
and smaller central banks in developing countries have been forced to implement
a range of adaptive measures, from tightening monetary policy to diversifying
their investment portfolios, to protect their economies from external shocks and
take advantage of available opportunities. The interaction between these global
and local economic strategies underscores the complex and interconnected nature
of the modern financial system, where decisions by a key player can have
widespread impacts on the rest of the world.

\xhdr{Node 1-2 (Level 2, NV-Embed-v2 similarity to root 0.9690)} In the constantly changing landscape of global economic governance, the recent
policy adjustments by the European Central Bank (ECB) are a significant sign of
the shifting tides in international finance. The ECB's decision to adjust its
Quantitative Easing (QE) program, along with a series of differentiated interest
rate adjustments, has sent shockwaves through financial markets, particularly
affecting emerging economies that are closely tied to the economic health of the
Eurozone. These changes, aimed at stabilizing the Euro and promoting economic
growth in the European Union, have had far-reaching consequences, including
capital flight, currency fluctuations, and an increase in risk aversion among
investors. The forward-looking communication of the ECB, intended to provide
clarity and predictability, has been met with mixed reactions, as some market
analysts argue that it lacks the necessary flexibility to adapt to the diverse
and dynamic economic conditions in different regions. In response, smaller
financial institutions and central banks of developing countries have been
forced to implement a variety of adaptive measures, from tightening monetary
policy to diversifying their investment portfolios, to protect their economies
from external shocks and capitalize on available opportunities. The interaction
between these global and local economic strategies underscores the complex and
interconnected nature of the modern financial system, where decisions by a major
player can have far-reaching impacts on the rest of the world.

\xhdr{Node 1-2-0 (Level 3, NV-Embed-v2 similarity to root 0.9738)} In the constantly evolving landscape of global economic governance, the recent
policy adjustments by the European Central Bank (ECB) are a significant sign of
changing tides in international finance. The ECB's decision to alter its
Quantitative Easing (QE) program, coupled with a series of differentiated
interest rate adjustments, has sent shockwaves through financial markets,
particularly affecting emerging economies closely tied to the economic health of
the eurozone. These changes, aimed at stabilizing the euro and promoting
economic growth in the European Union, have had significant consequences,
including capital outflows, currency fluctuations, and increased risk aversion
among investors. The ECB's forward guidance, intended to bring clarity and
predictability, has been met with mixed reactions, some market analysts arguing
that it lacks the necessary flexibility to adapt to the diverse and dynamic
economic conditions in different regions. In response, smaller financial
institutions and central banks in developing countries have been forced to
implement a variety of adaptive measures, ranging from monetary policy
tightening to diversification of their investment portfolios, to protect their
economies from external shocks and capitalize on available opportunities. The
interaction between these global and local economic strategies underscores the
complex and interconnected nature of the modern financial system, where
decisions by a major player can have significant impacts on the rest of the
world.

\xhdr{Node 1-2-1 (Level 3, NV-Embed-v2 similarity to root 0.9372)} In the constantly changing structure of global economic governance, the recent
political adjustments by the European Central Bank (ECB) are a clear sign of the
shifting tides in the international financial world. The ECB's decision to
adjust its Quantitative Easing (QE) program, along with a series of
differentiated interest rate adjustments, has sent shockwaves through financial
markets, particularly to emerging countries that are closely linked to the
economic health of the Eurozone. These changes, aimed at stabilizing the Euro
and promoting economic growth in the European Union, have had far-reaching
consequences, including capital flight, currency fluctuations, and increased
risk aversion among investors. The forward-looking communication of the ECB,
intended to provide clarity and predictability, has been met with mixed
reactions, as some market analysts argue that it lacks the necessary flexibility
to adapt to the diverse and dynamic economic conditions in different regions. In
response, smaller financial institutions and the central banks of developing
countries have been forced to implement a variety of adaptive measures, from
tightening monetary policy to diversifying their investment portfolios, to
protect their economies from external shocks and to benefit from available
opportunities. The interaction between these global and local economic
strategies underscores the complex and interconnected nature of the modern
financial system, where decisions by a major player can have far-reaching
impacts on the rest of the world.

\xhdr{Node 1-2-2 (Level 3, NV-Embed-v2 similarity to root 0.9695)} In the constantly changing landscape of global economic governance, recent
policy adjustments by the European Central Bank (ECB) are a significant sign of
the shifting tides in international finance. The ECB's decision to adjust its
Quantitative Easing (QE) program, along with a series of differential interest
rate adjustments, has sent shockwaves through financial markets, particularly
affecting emerging economies that are closely tied to the economic health of the
Eurozone. These changes, aimed at stabilizing the Euro and promoting economic
growth in the European Union, have had far-reaching consequences, including
capital flight, monetary fluctuations, and an increase in risk aversion among
investors. The forward-looking communication from the ECB, intended to provide
clarity and predictability, has been met with mixed reactions, as some market
analysts argue that it lacks the necessary flexibility to adapt to diverse and
dynamic economic conditions in different regions. In response, smaller financial
institutions and central banks of developing countries have been forced to
implement a variety of adaptive measures, from tightening monetary policy to
diversifying their investment portfolios, to protect their economies from
external shocks and capitalize on available opportunities. The interaction
between these global and local economic strategies underscores the complex and
interconnected nature of the modern financial system, where decisions by a major
player can have far-reaching impacts on the rest of the world.

\xhdr{Node 2 (Level 1, NV-Embed-v2 similarity to root 0.9723)} In the constantly evolving landscape of global economic governance, recent
policy adjustments by the European Central Bank (ECB) stand out as a significant
marker of changes in the currents of international finance. The ECB's decision
to recalibrate its quantitative easing (QE) program, along with a series of
subtle interest rate adjustments, has sent shockwaves through financial markets,
particularly affecting emerging economies that are intricately linked to the
economic health of the Eurozone. These changes, though designed to stabilize the
Euro and stimulate economic growth within the European Union, have had far-
reaching consequences, including capital flight, currency volatility, and
increased risk aversion among investors. The future orientation of the ECB,
which seeks to provide clarity and predictability, has been met with mixed
reactions, as some market analysts argue that it lacks the necessary flexibility
to address the diverse and dynamic economic conditions in different regions. In
response, smaller financial institutions and central banks of emerging markets
have been forced to adopt a series of adaptive measures, from tightening
monetary policies to diversifying their investment portfolios, in order to
protect their economies from external shocks and capitalize on emerging
opportunities. The interaction between these global and local economic
strategies underscores the complex and interconnected nature of the modern
financial system, where decisions by a major player can have profound
implications for the rest of the world.

\xhdr{Node 2-0 (Level 2, NV-Embed-v2 similarity to root 0.9709)} In the constantly evolving landscape of global economic governance, the recent
policy adjustments by the European Central Bank (ECB) stand out as a significant
marker of changes in the currents of international finance. The ECB's decision
to recalibrate its quantitative easing (QE) program, along with a series of
subtle interest rate adjustments, has sent shockwaves through financial markets,
particularly affecting emerging economies that are closely tied to the economic
health of the eurozone. These changes, although designed to stabilize the euro
and stimulate economic growth within the European Union, have had far-reaching
consequences, including capital flight, exchange rate volatility, and increased
risk aversion among investors. The future direction of the ECB, aimed at
providing clarity and predictability, has been met with mixed reactions, some
market analysts arguing that it lacks the necessary flexibility to respond to
the diverse and dynamic economic conditions in different regions. In response,
small financial institutions and central banks in emerging markets have been
forced to adopt a series of adaptive measures, ranging from easing monetary
policies to diversifying their investment portfolios, in order to protect their
economies from external shocks and capitalize on emerging opportunities. The
interaction between these global and local economic strategies underscores the
complex and interconnected nature of the modern financial system, where
decisions by a major player can have profound implications for the rest of the
world.

\xhdr{Node 2-0-0 (Level 3, NV-Embed-v2 similarity to root 0.9686)} In the constantly evolving landscape of global economic governance, the recent
policy adjustments by the European Central Bank (ECB) stand out as a significant
marker of changes in the currents of international finance. The ECB's decision
to recalibrate its quantitative easing (QE) program, coupled with a series of
subtle interest rate adjustments, has sent shockwaves through financial markets,
particularly affecting emerging economies that are closely tied to the economic
health of the eurozone. These changes, although designed to stabilize the euro
and boost economic growth within the European Union, have had broader
consequences, including capital flight, exchange rate volatility, and increased
risk aversion among investors. The future direction of the ECB, aimed at
providing clarity and predictability, has been met with mixed reactions, some
market analysts arguing that it lacks the necessary flexibility to respond to
the diverse and dynamic economic conditions in different regions. In response,
small financial institutions and central banks in emerging markets have been
forced to adopt a series of adaptation measures, ranging from easing monetary
policies to diversifying their investment portfolios, in order to protect their
economies from external shocks and capitalize on emerging opportunities. The
interaction between these global and local economic strategies underscores the
complex and interconnected nature of the modern financial system, where
decisions by a major player can have profound implications for the rest of the
world.

\xhdr{Node 2-0-1 (Level 3, NV-Embed-v2 similarity to root 0.9361)} In the constantly changing landscape of global economic governance, the recent
political adjustments by the European Central Bank (ECB) are a significant sign
of changes in the currents of international financial markets. The ECB's
decision to realign its Quantitative Easing (QE) program, coupled with a series
of subtle interest rate adjustments, has sent shockwaves through financial
markets, particularly in emerging countries that are closely tied to the
economic health of the Eurozone. These changes, although designed to stabilize
the Euro and promote economic growth within the European Union, have had far-
reaching consequences, including capital flight, exchange rate volatility, and
increased risk aversion among investors. The future direction of the ECB, which
aims for clarity and predictability, has been met with mixed reactions. Some
market analysts argue that it lacks the necessary flexibility to respond to the
diverse and dynamic economic conditions in different regions. In response, small
financial institutions and central banks in emerging markets have been forced to
adopt a range of adaptive measures, from loosening monetary policy to
diversifying their investment portfolios, to protect their economies from
external shocks and capitalize on new opportunities. The interaction between
these global and local economic strategies underscores the complex and
interconnected nature of the modern financial system, where decisions by a major
player can have far-reaching implications for the rest of the world.

\xhdr{Node 2-0-2 (Level 3, NV-Embed-v2 similarity to root 0.9701)} In the constantly evolving landscape of global economic governance, recent
policy adjustments by the European Central Bank (ECB) stand out as a significant
marker of changes in the currents of international finance. The ECB's decision
to recalibrate its quantitative easing (QE) program, along with a series of
subtle interest rate adjustments, has sent shockwaves through financial markets,
particularly affecting emerging economies that are closely tied to the economic
health of the eurozone. These changes, although designed to stabilize the euro
and stimulate economic growth within the European Union, have had far-reaching
consequences, including capital flight, exchange rate volatility, and an
increase in risk aversion among investors. The future direction of the ECB,
aimed at providing clarity and predictability, has been met with mixed
reactions, some market analysts arguing that it lacks the necessary flexibility
to respond to the various and dynamic economic conditions in different regions.
In response, small financial institutions and central banks in emerging markets
have been forced to adopt a series of adaptive measures, ranging from easing
monetary policies to diversifying their investment portfolios, in order to
protect their economies from external shocks and capitalize on emerging
opportunities. The interaction between these global and local economic
strategies underscores the complex and interconnected nature of the modern
financial system, where decisions by a major player can have profound
implications for the rest of the world.

\xhdr{Node 2-1 (Level 2, NV-Embed-v2 similarity to root 0.9395)} In the constantly changing landscape of global economic governance, the recent
political adjustments by the European Central Bank (ECB) stand out as a
significant sign of changes in the currents of international finance. The ECB's
decision to adjust its Quantitative Easing (QE) program, along with a series of
subtle interest rate adjustments, has sent shockwaves through the financial
markets, particularly the emerging economies that are closely tied to the
economic health of the Eurozone. These changes, although designed to stabilize
the euro and promote economic growth within the European Union, have had far-
reaching consequences, including capital flight, currency volatility, and
increased risk aversion among investors. The future orientation of the ECB,
which aims to provide clarity and predictability, has been met with mixed
reactions, as some market analysts argue that it lacks the necessary flexibility
to account for the diverse and dynamic economic conditions in different regions.
In response, smaller financial institutions and central banks of emerging
countries have been forced to adopt a series of adaptive measures, from
tightening monetary policy to diversifying their investment portfolios, to
protect their economies from external shocks and capitalize on emerging
opportunities. The interaction between these global and local economic
strategies underscores the complex and interconnected nature of the modern
financial system, where decisions by a major player can have far-reaching
implications for the rest of the world.

\xhdr{Node 2-1-0 (Level 3, NV-Embed-v2 similarity to root 0.9659)} In the constantly evolving landscape of global economic governance, the recent
policy adjustments by the European Central Bank (ECB) stand out as a significant
sign of changes in the currents of international finance. The ECB's decision to
adjust its Quantitative Easing (QE) program, along with a series of subtle
interest rate adjustments, has sent shockwaves through financial markets,
particularly emerging economies closely tied to the economic health of the
eurozone. These changes, although designed to stabilize the euro and promote
economic growth within the European Union, have had far-reaching consequences,
including capital flight, currency volatility, and increased risk aversion among
investors. The future direction of the ECB, which aims to provide clarity and
predictability, has been met with mixed reactions, some market analysts arguing
that it lacks the necessary flexibility to account for diverse and dynamic
economic conditions in different regions. In response, smaller financial
institutions and central banks of emerging countries have been forced to adopt a
series of adaptive measures, from tightening monetary policy to diversifying
their investment portfolios, to protect their economies from external shocks and
capitalize on emerging opportunities. The interaction between these global and
local economic strategies underscores the complex and interconnected nature of
the modern financial system, where the decisions of a major player can have far-
reaching implications for the rest of the world.

\xhdr{Node 2-1-1 (Level 3, NV-Embed-v2 similarity to root 0.9380)} In the constantly changing landscape of global economic governance, the recent
political adjustments by the European Central Bank (ECB) are a significant sign
of changes in the currents of international finance. The ECB's decision to
adjust its quantitative easing (QE) program, along with a series of subtle
interest rate adjustments, has sent shockwaves through financial markets,
particularly through the emerging economies that are closely tied to the
economic health of the Eurozone. These changes, although aimed at stabilizing
the euro and promoting economic growth within the European Union, have had far-
reaching consequences, including capital flight, currency fluctuations, and
increased risk aversion among investors. The future orientation of the ECB,
which is intended to provide clarity and predictability, has been met with mixed
reactions, as some market analysts argue that it lacks the necessary flexibility
to consider the diverse and dynamic economic conditions in different regions. In
response, smaller financial institutions and central banks of emerging countries
have been forced to adopt a range of adaptive measures, from tightening monetary
policy to diversifying their investment portfolios, to protect their economies
from external shocks and seize new opportunities. The interaction between these
global and local economic strategies underscores the complex and interconnected
nature of the modern financial system, where decisions by a major player can
have far-reaching implications for the rest of the world.

\xhdr{Node 2-1-2 (Level 3, NV-Embed-v2 similarity to root 0.9696)} In the constantly changing landscape of global economic governance, the recent
policy adjustments by the European Central Bank (ECB) stand out as a significant
sign of changes in the currents of international finance. The ECB's decision to
adjust its Quantitative Easing (QE) program, along with a series of subtle
interest rate adjustments, has sent shockwaves through financial markets,
particularly to emerging economies that are closely linked to the economic
health of the Eurozone. These changes, although designed to stabilize the euro
and promote economic growth within the European Union, have had far-reaching
consequences, including capital flight, currency volatility, and an increase in
risk aversion among investors. The future guidance of the ECB, which aims to
provide clarity and predictability, has been met with mixed reactions, as some
market analysts argue that it lacks the necessary flexibility to account for the
diverse and dynamic economic conditions in different regions. In response,
smaller financial institutions and central banks of emerging countries have been
forced to adopt a series of adaptive measures, from tightening monetary policy
to diversifying their investment portfolios, to protect their economies from
external shocks and take advantage of emerging opportunities. The interaction
between these global and local economic strategies underscores the complex and
intertwined nature of the modern financial system, where decisions by a major
player can have far-reaching implications for the rest of the world.

\xhdr{Node 2-2 (Level 2, NV-Embed-v2 similarity to root 0.9803)} In the constantly evolving landscape of global economic governance, recent
policy adjustments by the European Central Bank (ECB) stand out as a significant
marker of changes in the currents of international finance. The ECB's decision
to recalibrate its quantitative easing (QE) program, along with a series of
subtle interest rate adjustments, has sent shockwaves through financial markets,
particularly affecting emerging economies that are intricately linked to the
economic health of the Eurozone. These changes, although designed to stabilize
the Euro and stimulate economic growth within the European Union, have had far-
reaching consequences, including capital flight, currency volatility, and an
increase in risk aversion among investors. The future guidance of the ECB, which
aims to provide clarity and predictability, has been met with mixed reactions,
as some market analysts argue that it lacks the necessary flexibility to address
the diverse and dynamic economic conditions in different regions. In response,
smaller financial institutions and central banks of emerging markets have been
forced to adopt a series of adaptive measures, from tightening monetary policies
to diversifying their investment portfolios, in order to protect their economies
from external shocks and capitalize on emerging opportunities. The interaction
between these global and local economic strategies underscores the complex and
interconnected nature of the modern financial system, where decisions by a major
player can have profound implications for the rest of the world.

\xhdr{Node 2-2-0 (Level 3, NV-Embed-v2 similarity to root 0.9769)} In the constantly evolving landscape of global economic governance, the recent
policy adjustments by the European Central Bank (ECB) stand out as a significant
marker of changes in the currents of international finance. The ECB's decision
to recalibrate its quantitative easing (QE) program, combined with a series of
subtle interest rate adjustments, has sent shock waves through financial
markets, particularly affecting emerging economies that are closely tied to the
economic health of the eurozone. These changes, although designed to stabilize
the euro and boost economic growth within the European Union, have had
significant consequences, including capital flight, currency volatility, and
increased risk aversion among investors. The ECB's future guidance, aimed at
providing clarity and predictability, has been met with mixed reactions, some
market analysts arguing that it lacks the necessary flexibility to respond to
the diverse and dynamic economic conditions in different regions. In response,
small financial institutions and central banks in emerging markets have been
forced to adopt a series of adaptive measures, ranging from tightening monetary
policies to diversifying their investment portfolios, in order to protect their
economies from external shocks and capitalize on emerging opportunities. The
interaction between these global and local economic strategies underscores the
complex and interconnected nature of the modern financial system, where
decisions by a major player can have profound implications for the rest of the
world.

\xhdr{Node 2-2-1 (Level 3, NV-Embed-v2 similarity to root 0.9269)} In the constantly changing landscape of global economic governance, the recent
political adjustments by the European Central Bank (ECB) stand out as a
significant sign of changes in the flows of international finance. The ECB's
decision to recalibrate its Quantitative Easing (QE) program, along with a
series of subtle interest rate adjustments, has sent shockwaves through
financial markets, particularly affecting emerging economies that are closely
tied to the economic health of the Eurozone. These changes, although designed to
stabilize the euro and boost economic growth within the European Union, have had
far-reaching consequences, including capital flight, currency fluctuations, and
increased risk aversion among investors. The future leadership of the ECB, which
is supposed to provide clarity and predictability, has been met with mixed
reactions, as some market analysts argue that it lacks the necessary flexibility
to handle the diverse and dynamic economic conditions in different regions. In
response, smaller financial institutions and central banks of emerging markets
have been forced to adopt a range of adaptive measures, from tightening monetary
policy to diversifying their investment portfolios, to protect their economies
from external shocks and capitalize on new opportunities. The interaction
between these global and local economic strategies underscores the complex and
interconnected nature of the modern financial system, where decisions by a key
player can have far-reaching implications for the rest of the world.

\xhdr{Node 2-2-2 (Level 3, NV-Embed-v2 similarity to root 0.9806)} In the constantly evolving landscape of global economic governance, recent
policy adjustments by the European Central Bank (ECB) stand out as a significant
marker of changes in the currents of international finance. The ECB's decision
to recalibrate its quantitative easing (QE) program, along with a series of
subtle interest rate adjustments, has sent shockwaves through financial markets,
particularly affecting emerging economies that are intricately linked to the
economic health of the Eurozone. These changes, although designed to stabilize
the Euro and stimulate economic growth within the European Union, have had far-
reaching consequences, including capital flight, currency volatility, and an
increase in risk aversion among investors. The future guidance of the ECB, which
aims to provide clarity and predictability, has been met with mixed reactions,
as some market analysts argue that it lacks the necessary flexibility to address
the diverse and dynamic economic conditions in different regions. In response,
smaller financial institutions and central banks in emerging markets have been
forced to adopt a series of adaptive measures, from tightening monetary policies
to diversifying their investment portfolios, in order to protect their economies
from external shocks and capitalize on emerging opportunities. The interaction
between these global and local economic strategies underscores the complex and
interconnected nature of the modern financial system, where decisions by a major
player can have profound implications for the rest of the world.

}

\end{document}